\documentclass[11pt, a4paper]{article}

\usepackage[utf8]{inputenc}
\usepackage[T1]{fontenc}
\usepackage{lmodern}
\usepackage{microtype}

\usepackage{amsmath, amssymb}
\usepackage{graphicx}
\usepackage{booktabs}
\usepackage{hyperref}
\usepackage{cleveref}
\usepackage{makecell}
\usepackage[authoryear,longnamesfirst]{natbib}

\usepackage{subfig}

\usepackage[margin=2.5cm]{geometry}
\usepackage{tikz}
\usetikzlibrary{shapes.geometric, arrows.meta, positioning}

\title{A Modelling and Evaluation Framework for EuroCrops-Driven Sentinel-2 Crop Segmentation}

\author{Alexandra Nicoleta Scarlat \and Ioana Cristina Plajer \and Alexandra B\u{a}icoianu \\
\\
Transilvania University of Bra\c{s}ov, Faculty of Mathematics and Computer Science,\\ Bra\c{s}ov, Romania}

\date{}  

\begin{document}
\maketitle
\begin{abstract}
This work presents a configurable pipeline for generating semantic-segmentation-ready agricultural datasets from Sentinel-2 imagery and EuroCrops parcel-level annotations. The workflow transforms heterogeneous vector crop annotations into aligned multispectral image--mask pairs through label harmonization, Sentinel-2 product selection, spatial alignment, rasterization, patch extraction, quality filtering, and class-aware sample selection. The generated dataset contains 67,337 patches from five European countries and uses a reduced taxonomy of ten crop classes plus background.

A four-level U-Net with Group Normalization was trained using 10 Sentinel-2 spectral bands and a composite loss combining class-weighted cross-entropy and Dice loss. On the internal EuroCrops-based test split, the model achieved a mean Intersection over Union (mIoU) of 0.7665, a pixel accuracy of 0.8693, and a mean class accuracy of 0.9072. Compared with spectral and spatial-context Random Forest baselines, the U-Net showed the importance of learned multi-scale spatial representations for crop segmentation.

External evaluation was performed on unseen Belgian EuroCrops subsets, DACIA5, and PASTIS. The results show a clear performance gap under external and cross-dataset evaluation, especially for benchmarks with different taxonomies, annotation protocols, spatial coverage, or temporal organization. The model transfers more reliably to dominant and taxonomically aligned classes such as maize and wheat, while performance remains limited for several minority classes and for the adapted single-date PASTIS setting. These findings highlight both the potential and the limitations of using EuroCrops-derived supervision for Sentinel-2 crop segmentation under realistic domain shifts.
\end{abstract}


\section{Introduction}\label{sec:introduction}

As climate change, habitat loss, and global population pressures intensify, the sustainable management of agricultural land has become increasingly important. Accurate crop monitoring is now essential for estimating yields, tracking land-use shifts, and informing both agricultural practices and public policy. A key enabler of large-scale crop monitoring is multispectral (MS) imagery, acquired by sensors operating across a range of wavelengths spanning the visible and infrared domains, which provides substantially richer information about vegetation, crop development, and agricultural land use than conventional RGB imagery. Driven by the growing accessibility of such data through missions like the Copernicus Earth Observation Program, a constellation of satellites known as the Sentinels \citep{aschbacher2017esa}, and in particular the freely available 13-band MS data of Sentinel-2 \citep{esa2012sentinel2}, satellite-based crop mapping has emerged as a timely and significant research field, enabling continuous, large-scale observation of the world's food systems.
Within this context, semantic segmentation is particularly relevant, as it provides pixel-level land cover information that can support detailed agricultural mapping and downstream spatial applications. However, the task remains challenging due to the spectral and phenological similarity between crop types, the prevalence of imperfect labels, and the strong dependence of model performance on training data quality. Progress in this area is therefore closely conditioned on the availability of large-scale, well-annotated remote sensing datasets and benchmarks, whose development remains one of the central challenges in the field. 

One of the most widely used Sentinel-2 archives for remote sensing image understanding is BigEarthNet, a large collection of image patches annotated with multi-label land-cover information \citep{bigearthnet}. Although it has contributed substantially to the development of deep learning methods for satellite image analysis, its task formulation targets image-level land-cover classification rather than pixel-level crop segmentation.

More specialized datasets have been proposed for crop-type mapping from Sentinel-2 data, such as TimeSen2Crop, which provides over one million labeled Sentinel-2 time-series samples associated with crop types, making it well-suited for temporal crop classification tasks \citep{timesen2crop}. However, its formulation is based on labeled temporal samples rather than aligned image-mask pairs, and is therefore not directly applicable to semantic segmentation.

AgriFieldNet India is another relevant dataset for crop type mapping from Sentinel-2 imagery. It provides MS Sentinel-2 observations organized into $256 \times 256$ chips, together with crop-type labels derived from ground survey data \citep{agrifieldnet}. This makes it closer to patch-based crop mapping than purely sample-based time-series datasets, although its benchmark setting and geographical context differ from the segmentation workflow proposed in this work.

At European scale, AgriSen-COG introduces a large-scale, multicountry and multitemporal Sentinel-2 benchmark for crop mapping using LPIS-based ground truth \citep{agrisencog}. It is particularly relevant because it addresses challenges also encountered in this work, including geographic domain shift, crop-label heterogeneity, pixel-level crop mapping, and parcel-based agricultural annotations. Sen4AgriNet is also closely related, as it provides a Sentinel-2 multi-year and multi-country benchmark for crop classification and segmentation using annotations derived from Land Parcel Identification System data \citep{sen4agrinet}. Both datasets demonstrate the importance of harmonized agricultural benchmarks, while the present work focuses specifically on a configurable workflow for generating semantic-segmentation-ready data from EuroCrops and Sentinel-2 according to a user-defined taxonomy and sampling strategy.

The Panoptic Agricultural Satellite Time Series (PASTIS) \citep{garnot2021pastis,pastis_dataset} provides a benchmark for semantic and panoptic segmentation of agricultural parcels from Sentinel-2 image time series. It is closely related to this work because it provides pixel-level agricultural parcel annotations and is explicitly designed for segmentation. However, PASTIS is a relatively small dataset, built around temporal Sentinel-2 sequences, and geographically restricted. In our study,  PASTIS is used as an external cross-protocol evaluation benchmark.

In terms of modelling, crop mapping has traditionally relied on classical machine learning algorithms such as Random Forests (RF). These remain robust baselines in remote sensing because of their ability to handle high-dimensional spectral data and complex, non-linear relationships between input features and target labels \citep{breiman2001random,belgiu2016random}. More recently, crop mapping has increasingly been formulated as a dense prediction or semantic segmentation problem, where each pixel is assigned to a crop or land-cover class. In this setting, deep learning models such as Convolutional Neural Networks (CNNs), Fully Convolutional Networks (FCNs), and U-Net-like architectures are particularly relevant because they can learn spatial context directly from multispectral image patches rather than relying only on handcrafted pixel-level features \citep{gao2023crop_mapping_segmentation,li2023crop_segmentation}. Within this domain, the U-Net architecture is especially suitable; its symmetric encoder--decoder structure and skip connections allow the model to preserve high-resolution spatial details while capturing deeper semantic context, making it useful for delineating field boundaries and parcel-level crop structures \citep{ronneberger2015unet}.

Despite recent advances, many existing Earth Observation (EO) resources are distributed with rigid taxonomies, fixed spatial tiling, or task-specific preprocessing assumptions. These constraints often limit the flexibility required to construct custom semantic segmentation datasets with precise class definitions, tailored patch selection criteria, and rigorous image-mask alignment.

To address these limitations, we propose a reliable pipeline for generating high-quality training data by pairing Sentinel-2 multispectral imagery with EuroCrops vector-based parcel annotations, through a process of careful spatial alignment, rasterization, and patch extraction. Furthermore, quality-aware sample filtering is performed under realistic quality constraints, addressing one of the primary challenges in remote sensing: ensuring the consistency of ground-truth labels with respect to dynamic satellite observations.

The main contributions of this work can be summarized as follows:
\begin{itemize}
\item A configurable data generation pipeline integrating Sentinel-2 multispectral imagery with EuroCrops parcel-level annotations to produce semantic-segmentation-ready agricultural datasets, supporting adaptable class taxonomies, controlled patch selection, and scalable generation of aligned image--mask pairs;
\item A reduced and harmonized crop taxonomy focused on major European agricultural classes, designed for compatibility with future Romanian applications;
\item A quality-aware sample filtering strategy under realistic operational constraints, ensuring the consistency of ground-truth labels with respect to dynamic satellite observations;
\item A class-aware patch selection strategy to mitigate class imbalance arising from background and frequently occurring crop classes;
\item A deep learning baseline in the form of a four-level U-Net, trained and evaluated on the generated dataset, alongside Random Forest baselines using spectral and local spatial-context features, enabling analysis of the contribution of spectral information, handcrafted local context, and learned spatial representations to the crop segmentation task;
\item An external evaluation protocol assessing model generalization across unseen Belgian EuroCrops subsets, the Romanian DACIA5 benchmark, and the PASTIS dataset, covering diverse geographic, taxonomic, and data-organization conditions.
\end{itemize}

The remainder of this paper is organized as follows. Section~2 describes the construction of the proposed EuroCrops-based Sentinel-2 dataset, including the input data sources, target taxonomy, label harmonization, data generation pipeline, and final dataset composition. Section~3 presents the experimental setup, including the U-Net model, Random Forest baselines, training configuration, evaluation metrics, and external evaluation datasets. Section~4 reports and discusses the internal and external experimental results. Finally, Section~5 summarizes the main conclusions, discusses the limitations of the study, and outlines possible directions for future work.

\section{Dataset Construction}\label{sec:materials_and_methods}

The dataset generated in this study combines Sentinel-2 Level-2A imagery with EuroCrops \citep{eurocrops} parcel annotations from five European nations: France, Germany, Austria, Slovakia, and the Czech Republic. These countries were selected to ensure both spatial diversity and a sufficient sample distribution for our target crop classes. This geographic distribution establishes a robust baseline for evaluating crop mapping performance across Europe, and specifically for testing generalization to Romania.

\subsection{Input data sources}\label{subsec:source_datasets}

Sentinel-2 is a multispectral Earth observation mission designed for systematic land monitoring applications, with strong relevance for agricultural analysis. Its MultiSpectral Instrument (MSI) acquires imagery in 13 spectral bands distributed at spatial resolutions of 10~m, 20~m, and 60~m, with a revisit time of approximately five days \citep{esa2012sentinel2,sentiwiki_s2}. These characteristics make Sentinel-2 particularly suitable for crop mapping, as they allow the observation of agricultural parcels at sufficient spatial detail and with repeated temporal coverage throughout the vegetation season.

In this work, Sentinel-2 Level-2A products were used as the main source of multispectral imagery. Beyond their spectral richness, these products are particularly useful because they include a Scene Classification Layer (SCL), which can support cloud masking and patch-level quality filtering during preprocessing \citep{sentinel2_l2a_docs}. The combination of visible, near-infrared, red-edge, and short-wave infrared bands makes Sentinel-2 highly relevant for distinguishing crop-related spectral patterns, vegetation vigour, and seasonal variability.

EuroCrops is a large-scale, parcel-level crop dataset derived from self-declared agricultural field annotations collected under the European Union's Common Agricultural Policy (CAP) \citep{eurocrops}. It provides vector-based parcel geometries with crop labels from multiple European countries, making it a valuable reference source for agricultural mapping. However, EuroCrops is not directly suitable for semantic segmentation in its raw form, as it presents inconsistencies in crop naming conventions and metadata across national sources, and provides annotations as vector geometries rather than pixel-level masks. Its use in a segmentation framework therefore requires label harmonization, spatial alignment with Sentinel-2 imagery, rasterization, and patch generation.
In this study, EuroCrops serves as the primary annotation source, integrating multiple country-year subsets from  France 2018, Austria 2021, Slovakia 2021, the Czech Republic 2023, and several German regional subsets from 2021 and 2023, increasing spatial and temporal diversity while requiring careful preprocessing to ensure consistency. Due to limited availability of annotated parcel-level data for Romania, the training dataset was constructed from other European subsets; however, the target taxonomy was defined with particular attention to crop categories relevant to the Romanian agricultural context, where the approach is intended to be applied and externally evaluated.

\subsection{Target taxonomy and class harmonization}\label{subsec:target_taxonomy_and_class_harmonization}

To enable efficient model training, the original EuroCrops labels were mapped to a reduced and standardized set of target classes. This step was necessary because EuroCrops integrates annotations from multiple national sources, where crop names may differ in language, spelling, specificity, and terminology. The selected classes correspond to representative crop categories commonly found in European agricultural landscapes, while also preserving compatibility with the Romanian agricultural context considered in the external DACIA5 \citep{dacia5paper,dacia5dataset} dataset evaluation.

The final taxonomy was designed to cover major cereal crops, industrial crops, root crops, legumes, and grassland-related agricultural areas, while maintaining a tractable classification problem for semantic segmentation. The selected target classes are maize, rapeseed, wheat, barley, oats, sunflower, sugar beet, soybean, potato, and pasture. In addition, a background class was used to represent non-target, undefined, or discarded areas.

The target taxonomy was implemented as a configurable rule-based mapping between heterogeneous EuroCrops crop names and the reduced set of semantic segmentation classes. Each target class was assigned a numerical identifier used in the generated masks. A parcel label was assigned to a target class if it matched at least one inclusion keyword and did not match any exclusion keyword associated with that class. Inclusion keywords were used to capture naming variants found across national EuroCrops datasets, while exclusion keywords were used as safeguards against ambiguous partial matches or semantically different crops with similar names. For example, \textit{grape} was excluded from the rapeseed class because it may match the substring "rape", while \textit{buckwheat} was excluded from the wheat class because it contains the word "wheat" but does not belong to the intended cereal grouping.

This rule-based mapping represents a compromise between agronomic relevance, label availability, compatibility with the Romanian application context, and the need to maintain a manageable number of classes for semantic segmentation. The resulting taxonomy is summarized in Table~\ref{tab:target_taxonomy}.

\begin{table}[t]
\caption{Target crop taxonomy and rule-based label mapping used for mask generation.}
\label{tab:target_taxonomy}
\small
\begin{tabular*}{\textwidth}{@{\extracolsep{\fill}} c l l l @{}}
\toprule
Class ID & Target class & Inclusion keywords & Exclusion keywords \\
\midrule
0 & Background & Non-target, undefined, or discarded areas & -- \\
1 & Maize & maize, corn & -- \\
2 & Rapeseed & rapeseed, rape seed, winter rape, oilseed rape & grape \\
3 & Wheat & wheat, triticale, rye & buckwheat \\
4 & Barley & barley & -- \\
5 & Oats & oats & -- \\
6 & Sunflower & sunflower & -- \\
7 & Sugar beet & sugar beet, beet & fodder beet, beetroot \\
8 & Soybean & soybean, soy beans, soy & -- \\
9 & Potato & potato & -- \\
10 & Pasture & pasture, grassland, meadow, arable grass, clover, \\
& & lucerne, alfalfa & -- \\
\bottomrule
\end{tabular*}
\end{table}
 
As shown in Table~\ref{tab:target_taxonomy}, the taxonomy includes ten agricultural target classes and one background class. Some target classes required aggregation in order to reduce ambiguity and improve label consistency. In particular, wheat, rye, and triticale were grouped into a single wheat class because they are agronomically related cereal crops and present similar spectral and phenological characteristics in single-date Sentinel-2 imagery. Similarly, the pasture class aggregates grassland and forage-related categories, including meadow, clover, lucerne, and alfalfa, because these labels represent related vegetated agricultural surfaces rather than clearly separable crop types in the selected setup.

Backed by Eurostat data \citep{eurostat}, the grouping of rye and triticale into the broader wheat category is justified by both agronomic similarities and European cultivation statistics. At the European Union level, rye and triticale occupy a substantially smaller cultivated area than dominant cereals like wheat and barley. Because they are agronomically and spectrally similar to wheat, separating them is exceptionally difficult in a single-date multispectral segmentation setting. Merging them ensures a more balanced and robust target category while reflecting the wider European agricultural context.

For the remaining categories, oats, sunflower, sugar beet, and soybean were designated as rare classes within the dataset configuration. These categories were explicitly prioritized during a class-aware patch selection process. This targeted sampling strategy was essential to preserve sufficient samples of underrepresented crops, thereby preventing dominant, highly frequent classes from overwhelming the model during dataset construction.

To transform the harmonized parcel-level annotations into a suitable dataset for model training, a dedicated data generation pipeline was designed. The following section describes the processing steps required to generate aligned image-mask pairs from Sentinel-2 imagery and EuroCrops data.

\subsection{Data generation pipeline}\label{subsec:data_generation_pipeline}

\begin{figure}[]
\centering
\includegraphics[width = \textwidth]{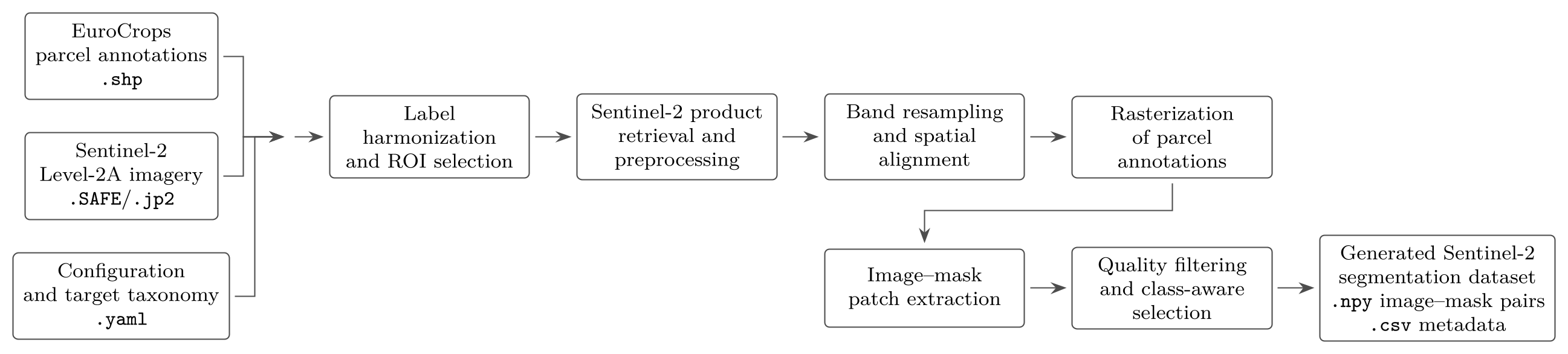}

\caption{Overview of the proposed Sentinel-2 and EuroCrops data generation workflow. The pipeline transforms EuroCrops parcel annotations, Sentinel-2 Level-2A imagery, and configuration files into aligned NumPy image--mask pairs and patch-level metadata for semantic segmentation.}
\label{fig:data_generation_pipeline}
\end{figure}

The pipeline, illustrated in Figure \ref{fig:data_generation_pipeline}, is structured as a coherent sequence of stages comprising label harmonization, region selection, satellite data retrieval, preprocessing, and patch extraction, each building on the previous to produce aligned image-mask pairs suitable for model training.

As a first step, the EuroCrops parcel annotations were filtered and mapped to a reduced set of target crop classes. Since EuroCrops contains heterogeneous labels originating from multiple national sources, crop names were harmonized through a rule-based mapping scheme, and parcels that could not be assigned to one of the selected target classes were discarded. This step reduced label ambiguity and ensured that the final masks only contained the classes relevant for the segmentation task.

To avoid processing large areas with little or no useful agricultural content, the study area was divided into candidate regions of interest (ROIs). Two ROI selection strategies were supported: a grid-based coverage strategy and a hot-cell strategy for target-class-focused sampling. In the grid-based strategy, candidate regions were ranked according to the amount of crop-covered area they contained, while in the hot-cell strategy, regions were centred around areas with high concentration of a target crop class. Only the selected ROIs were forwarded to the Sentinel-2 search stage.

For each ROI, Sentinel-2 Level-2A products were queried through the Copernicus Data Space Ecosystem STAC API using spatial, temporal, and cloud-cover constraints \citep{copernicus_stac_api}. Products whose footprints did not intersect the EuroCrops polygons were discarded. The remaining products were downloaded as SAFE archives and cached locally in order to avoid repeated downloads in subsequent runs. This design allowed dataset generation to be incremental while keeping the processing state associated with each ROI-product pair.

The Sentinel-2 products were selected within a consistent seasonal window spanning from April to September, corresponding to the main vegetation period. While the acquisition year was adapted to each EuroCrops subset according to data availability, the same monthly interval was preserved across all datasets.
This strategy ensures seasonal consistency while allowing the pipeline to make use of the most relevant annotation year available for each country. By restricting the search to the active growing season, the dataset captures meaningful crop-specific spectral responses and reduces the inclusion of non-informative observations such as bare soil or post-harvest fields.

After a valid product was retrieved, the spectral bands required by the dataset configuration were loaded from the SAFE structure. All bands were resampled and aligned to a common 10 m reference grid using band B02 as spatial reference. This ensured that all extracted image patches shared the same coordinate system, pixel spacing and extent. In parallel, the EuroCrops parcel geometries intersecting the ROI were reprojected to the Sentinel-2 image reference system and rasterized to produce pixel-level masks aligned with the multispectral image subset.

\begin{figure}[htbp]
    \centering
    \begin{minipage}[t]{0.78\linewidth}
        \vspace{0pt}
        \centering
        \includegraphics[height=5.5cm]{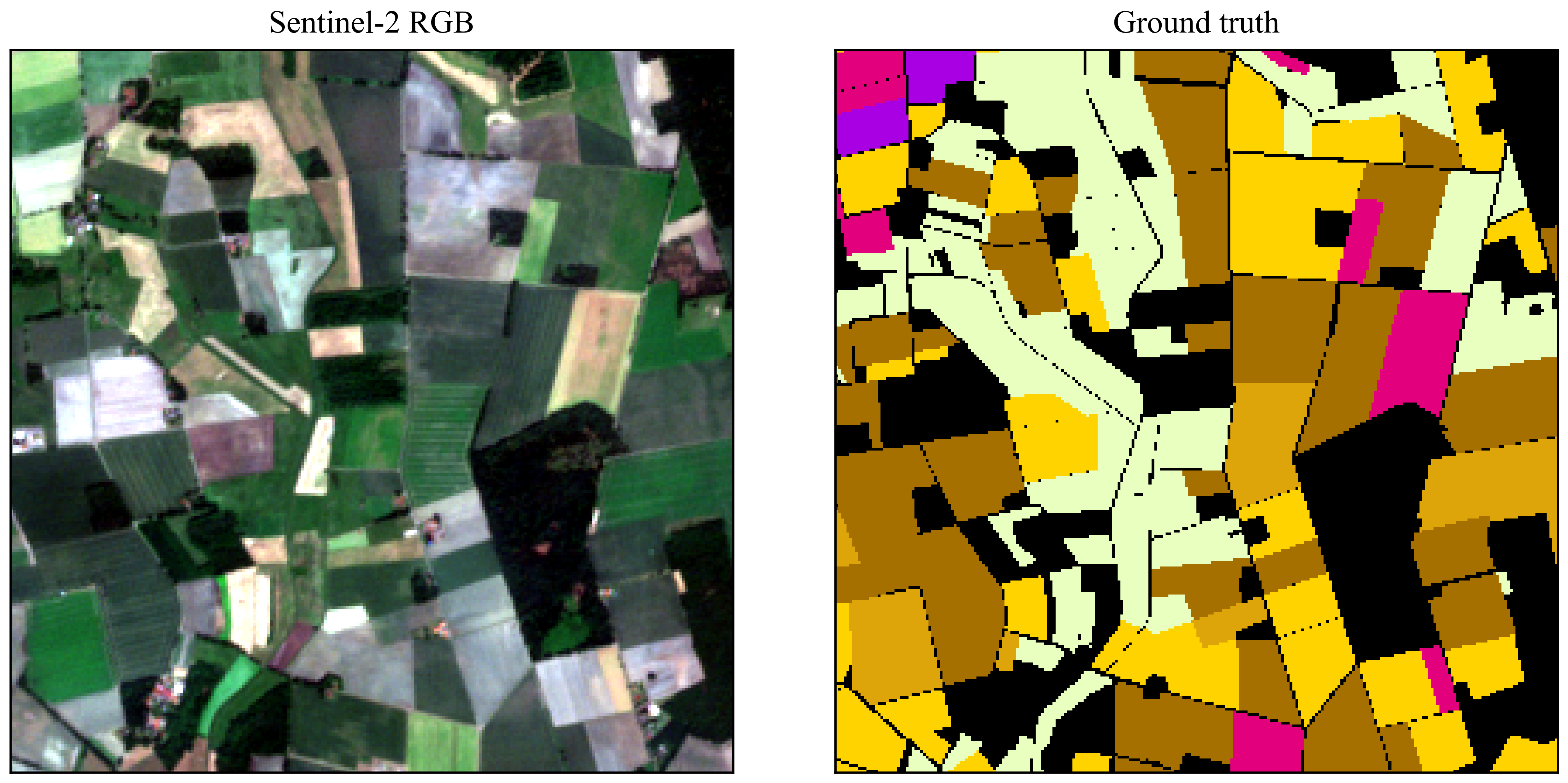}
    \end{minipage}
    \hspace{0.01\linewidth}
    \begin{minipage}[t]{0.16\linewidth}
        \vspace{0pt}
        \centering
        \includegraphics[height=5.5cm]{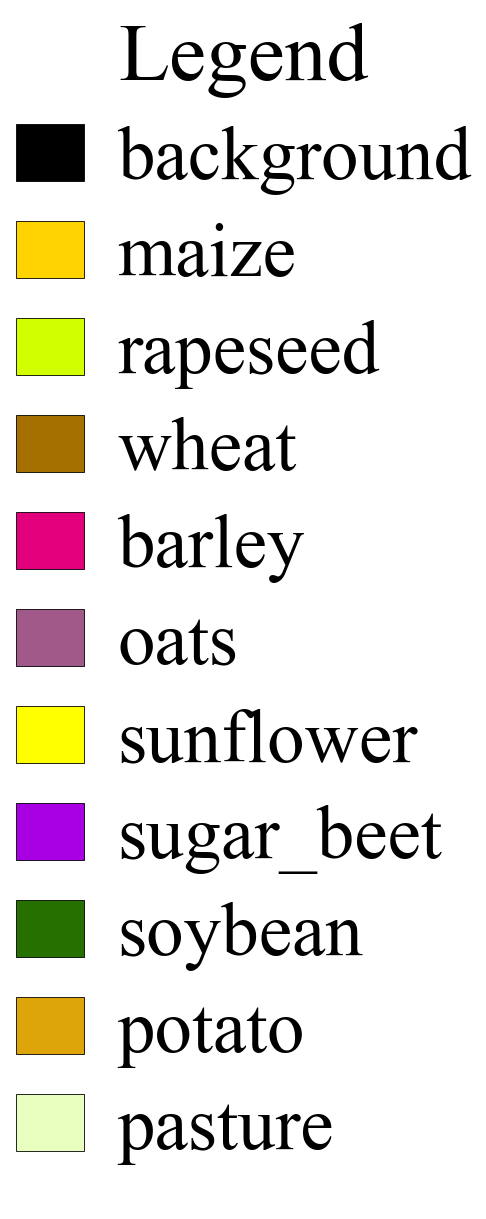}
    \end{minipage}
    \caption{Example of an image--mask pair generated by the proposed pipeline. Left: RGB visualization of a Sentinel-2 patch and the corresponding segmentation mask obtained by rasterizing the EuroCrops parcel annotations. Right: class-colour legend used for mask visualization, aligned with the EUCROPMAP colour convention.}
    \label{fig:rgb_mask}
\end{figure}

Patch extraction was then performed using a sliding-window strategy with a fixed patch size and stride. For each ROI, the pipeline generated candidate patches and applied a sequence of quality filters. Patches were rejected if they contained too much no-data area, insufficient foreground coverage, invalid or cloud-contaminated pixels according to the Scene Classification Layer (SCL), or unusually bright pixels likely associated with residual clouds or artifacts. In addition, the pipeline supported a target-class mode in which only patches containing a minimum amount of a specified crop class were retained, enabling focused generation for rare classes.
Figure \ref{fig:rgb_mask} illustrates an example image-mask pair generated by the pipeline, with the legend indicating the colour palette assigned to each crop class. For visual consistency, the colour encoding used in the qualitative mask and prediction visualizations was aligned with the EUCROPMAP class-colour convention \citep{eucropmap_gee}. This choice affects only the visualization of masks and predictions; the generated masks used for training and evaluation remain numerically encoded according to the project class identifiers defined in Table~\ref{tab:target_taxonomy}.

The dataset was built using 10 Sentinel-2 spectral bands selected for their relevance to agricultural monitoring and crop discrimination. Specifically, the input representation includes the visible bands B02, B03, and B04, the near-infrared bands B08 and B8A, the red-edge bands B05, B06, and B07, and the short-wave infrared bands B11 and B12. Detailed specifications for these bands can be found in official ESA Sentinel-2 User Handbook \citep{ESA2015Sentinel2}.

The output of this stage consists of aligned multispectral image patches and their corresponding segmentation masks, saved in NumPy \texttt{.npy} format, together with patch-level metadata describing the source product, spatial location, acquisition date, quality indicators, and class composition. These generated samples form the candidate pool from which the final dataset manifest was constructed.

After applying the preprocessing and quality filtering stages of the proposed pipeline, a total of 97,945 unique candidate patches were generated across all selected EuroCrops subsets. These patches were obtained after spatial alignment, rasterization of parcel annotations, Sentinel-2 band loading and resampling, cloud and invalid-pixel filtering, and patch-level quality constraints.
From this initial collection, a final subset of 67,337 patches was retained for model training, validation, and testing. The selection was not purely random, but guided by quality and class-awareness criteria. The objective was to reduce redundancy, limit the dominance of background and highly frequent crop classes, and preserve samples containing underrepresented categories whenever possible. This resulted in a more informative dataset while maintaining realistic agricultural variability.

\subsection{Final Dataset Overview}

The final dataset includes samples collected from five European countries: Austria, Germany, France, Slovakia, and the Czech Republic. The distribution of the patches per country is presented in Table \ref{tab:patches_by_country}.

\begin{table}[]
\caption{Distribution of patches by country in the final EuroCrops-based dataset.}
\label{tab:patches_by_country}
\begin{tabular*}{\textwidth}{@{\extracolsep{\fill}} l r r @{}}
\toprule
Country & Number of patches & Percentage \\
\midrule
Austria & 29,774 & 44.216\% \\
Germany & 13,307 & 19.762\% \\
France & 9,028 & 13.407\% \\
Slovakia & 8,358 & 12.412\% \\
Czech Republic & 6,870 & 10.202\% \\
\midrule
Total & 67,337 & 100.000\% \\
\bottomrule
\end{tabular*}
\end{table}

The larger contribution of Austria is mainly explained by the inclusion of several class-focused subsets generated to improve the representation of underrepresented crops such as oats, sunflower, soybean, and sugar beet, as the dataset was not intended to be geographically balanced across countries. Instead, the final manifest was constructed with the objective of obtaining sufficient class coverage for semantic segmentation, while still preserving spatial diversity across multiple European regions. 

To further analyze the composition of the final manifest, the pixel-level distribution was computed over all 67,337 retained patches. In total, the final dataset contains 4,412,997,632 labeled pixels. The distribution remains imbalanced, which is expected in agricultural segmentation datasets, but the class-aware selection strategy reduces the excessive dominance of the most frequent classes and preserves underrepresented crop categories.

\begin{table}[]
\caption{Pixel-level class distribution and patch-based analysis in the final EuroCrops-based dataset.}
\label{tab:class_distribution}
\begin{tabular*}{\textwidth}{@{} p{2.0cm} >{\raggedleft\arraybackslash}p{2.5cm} >{\raggedleft\arraybackslash}p{2.5cm} | p{2.0cm} >{\raggedleft\arraybackslash}p{2.5cm} >{\raggedleft\arraybackslash}p{2.5cm} @{}}
\toprule
Class & \makecell[r]{Number \\ of pixels} & \makecell[r]{Pixel \\ percentage} & Class & \makecell[r]{Number \\ of patches} & \makecell[r]{Patch \\percentage} \\
\midrule
Background  & 1,400,769,381 & 31.742\% & Background  & 67,337 & 100.000\% \\
Wheat       & 1,031,982,204 & 23.385\% & Wheat       & 66,939 & 99.409\%  \\
Maize       & 484,866,662   & 10.987\% & Barley      & 64,568 & 95.888\%  \\
Barley      & 373,327,263   & 8.460\%  & Maize       & 63,526 & 94.340\%  \\
Pasture     & 328,714,521   & 7.449\%  & Pasture     & 59,807 & 88.817\%  \\
Rapeseed    & 254,469,439   & 5.766\%  & Rapeseed    & 53,284 & 79.130\%  \\
Sugar beet  & 180,744,275   & 4.096\%  & Potato      & 41,103 & 61.041\%  \\
Soybean     & 108,360,177   & 2.455\%  & Sugar beet  & 38,593 & 57.313\%  \\
Potato      & 96,355,478    & 2.183\%  & Oats        & 27,221 & 40.425\%  \\
Sunflower   & 93,413,035    & 2.117\%  & Sunflower   & 27,087 & 40.226\%  \\
Oats        & 59,995,197    & 1.360\%  & Soybean     & 23,771 & 35.302\%  \\
\bottomrule
\end{tabular*}
\end{table}

As shown in columns 2 and 3 of Table~\ref{tab:class_distribution}, background and wheat remain the most represented classes, accounting for 31.742\% and 23.385\% of all labeled pixels, respectively. Maize and barley also have substantial representation, while oats, sunflower, soybean, and potato remain less frequent. This confirms that class imbalance is still present, but the final dataset includes all target classes with sufficient pixel-level support for training and evaluation.

In addition to pixel-level distribution, class presence was computed at patch level. This measures how many patches contain at least one pixel of each class, regardless of the area occupied by that class inside the patch. This analysis is useful because some classes may occupy a small pixel percentage while still appearing in a large number of patches.

The patch-level presence analysis (Table \ref{tab:class_distribution}, last two columns) shows that most patches contain multiple crop categories. Wheat, barley, maize, pasture, and rapeseed appear in a large proportion of the dataset, while soybean, sunflower, and oats are present in fewer patches. This behaviour is consistent with the pixel-level distribution and highlights the need for class-aware sampling and class-weighted training strategies.

The proposed pipeline is designed to be modular and configurable, enabling adaptation of both class definitions and sampling strategies. By decoupling label mapping from the core processing logic, the framework supports experimentation with different class groupings and target taxonomies.

Furthermore, patch generation is constrained at multiple levels, including per-region and per-product limits, in order to avoid dataset redundancy and to maintain a balanced distribution of samples. This controlled generation process improves both scalability and robustness of the resulting dataset.

\section{Experimental Setup}\label{subsec:models_and_training_procedure}
This section describes the U-Net segmentation model proposed for crop semantic segmentation, alongside two Random Forest baselines used for comparison. The architecture, hyperparameters, and training configuration are detailed for each model. The evaluation metrics used throughout the experiments are then defined, followed by a brief description of the external datasets considered for generalization testing.

The dataset constructed based on the pipeline described in the preceding section was divided into training, validation, and test subsets using an approximately 80\%, 10\%, and 10\% split. This resulted in 53,888 patches for training, 6,797 patches for validation, and 6,652 patches for testing.

For the EuroCrops-based dataset, the Sentinel-2 Level-2A values were normalized by dividing each selected band by 10000. The model used the same 10-band input order for all datasets: B02, B03, B04, B08, B05, B06, B07, B8A, B11, and B12. These bands include visible, red-edge, near-infrared, and short-wave infrared information, which is useful for distinguishing crop types.

\subsection{U-Net segmentation model}\label{subsubsec:unet_segmentation_model}

U-Net is an encoder-decoder convolutional architecture originally proposed for image segmentation \citep{ronneberger2015unet}. Its skip connections combine high-level semantic features with low-level spatial details, making it suitable for dense prediction tasks where both class discrimination and boundary preservation are important.

These properties make U-Net particularly suitable for crop segmentation from multispectral satellite imagery, where both semantic discrimination and spatial precision are important. In agricultural scenes, the model must distinguish between spectrally similar crop types while also preserving parcel boundaries and local field structure.

The depth of our network, was limited to four resolution levels as a trade-off between model capacity and computational efficiency. Given the input patch size of $256 \times 256$, this configuration allows sufficient receptive field to capture contextual information while avoiding excessive downsampling that could lead to loss of fine spatial details. At the same time, it ensures feasible training under hardware constraints.

The model was trained using a composite loss function combining class-weighted cross-entropy with a Dice-based loss component. Cross-entropy provides stable pixel-wise supervision for multi-class semantic segmentation, while the Dice component encourages better region-level overlap between predicted and reference masks and has been widely used for segmentation problems affected by class imbalance \citep{milletari2016vnet,sudre2017generalised}.

Because the generated dataset remains imbalanced at pixel level, a weighted version of cross-entropy was used. Lower weights were assigned to dominant classes such as background and wheat, while higher weights were assigned to underrepresented classes such as oats, sunflower, soybean, and potato. This weighting strategy was intended to reduce the dominance of frequent classes during optimization and to increase the contribution of minority crop categories and the respective weights for each class are presented in Table \ref{tab:class_weights}.

The final loss function is defined in Eq. \ref{eq:loss} as:
\begin{equation}
\mathcal{L} = (1 - \lambda)\,\mathcal{L}_{WCE} + \lambda\,\mathcal{L}_{Dice}
\label{eq:loss}
\end{equation}
In Eq. \ref{eq:loss} $\mathcal{L}_{WCE}$ denotes the class-weighted cross-entropy loss, $\mathcal{L}_{Dice}$ denotes the Dice loss, and $\lambda = 0.4$ controls the contribution of the Dice component. The Dice loss was computed while excluding the background class, so that the overlap component focused on agricultural crop regions rather than on the dominant background class.

\begin{table}[]
\caption{Class weights used in the weighted cross-entropy component of the U-Net loss.}
\label{tab:class_weights}
\begin{tabular*}{\textwidth}{@{}p{4.0cm} p{3.0cm}@{}}
\toprule
Class & Weight \\
\midrule
Background & 0.25 \\
Maize & 1.00 \\
Rapeseed & 1.50 \\
Wheat & 0.75 \\
Barley & 1.00 \\
Oats & 3.00 \\
Sunflower & 2.50 \\
Sugar beet & 1.50 \\
Soybean & 2.00 \\
Potato & 3.00 \\
Pasture & 1.00 \\
\bottomrule
\end{tabular*}
\end{table}

\subsubsection*{Data augmentation}\label{par:data_augmentation}
To improve generalization while preserving the radiometric consistency of the input data, a conservative data augmentation strategy was applied exclusively to the training set. This approach deliberately omitted spectral perturbations, as multispectral Sentinel-2 channels carry physically meaningful reflectance information that must remain intact. Instead, spatial variability was increased through random geometric transformations, specifically horizontal flips, vertical flips, and rotations by multiples of 90 degrees. These spatial transformations were applied jointly to both the multispectral image patch and its corresponding segmentation mask to strictly preserve pixel-wise alignment.
 
\subsubsection*{Training setup}\label{par:training_setup}

The architecture was implemented as a four-level U-Net with Group Normalization (GN), selected over Batch Normalization because hardware constraints imposed a small batch size, a setting that typically destabilizes batch-dependent normalization \citep{wu2018groupnorm}. The model was trained using the Adam optimizer \citep{kingma2015adam} with an initial learning rate of $3 \times 10^{-4}$, reduced by a factor of 0.5 upon validation plateau, over a maximum of 260 epochs. The checkpoint selected for testing corresponds to epoch 179, chosen according to the validation loss criterion. The highest validation mIoU of 0.7639 was observed at epoch 195; however, the difference relative to epoch 179 was negligible, indicating that the model had already reached a stable performance plateau. The full training configuration is summarized in Table~\ref{tab:training_config}.

\begin{table}[]
\centering
\caption{Main training configuration for the proposed U-Net model.}
\begin{tabular*}{\textwidth}{@{} l l @{}}
\toprule
Parameter & Value \\
\midrule
\multicolumn{2}{l}{\textit{Architecture}} \\
\quad Model class        & UNet4LevelsGN \\
\quad Levels             & 4 \\
\quad Input bands        & 10 Sentinel-2 bands \\
\quad Input patch size   & $256 \times 256$ \\
\quad Number of classes  & 11 \\
\quad Base channels      & 48 \\
\quad Normalization      & Group Normalization \\
\quad Upsampling mode    & Bilinear \\
\quad Bottleneck dropout & 0.1 \\
\quad Decoder dropout    & 0.0 \\
\quad Trainable parameters & 17,660,027 \\
\midrule
\multicolumn{2}{l}{\textit{Training}} \\
\quad Batch size         & 2 \\
\quad Optimizer          & Adam \\
\quad Initial learning rate & $3 \times 10^{-4}$ \\
\quad LR reduction factor & 0.5 \\
\quad Maximum epochs     & 260 \\
\quad Best checkpoint epoch & 179 \\
\quad Checkpoint size    & 70.66 MB \\
\midrule
\multicolumn{2}{l}{\textit{Loss function}} \\
\quad Cross-entropy weighting & Class-weighted \\
\quad Dice weight        & 0.4 \\
\quad Dice background    & Excluded \\
\quad Ignored label      & None \\
\bottomrule
\end{tabular*}
\label{tab:training_config}
\end{table}

The preparatory workflow, which included dataset organization, preprocessing checks, visualization, and auxiliary scripting, was carried out on a local laptop. Model training was conducted on a high-performance server kindly provided by the Transilvania University of Bra\c sov. The detailed specifications of this infrastructure are provided in Table~\ref{tab:computational_environment}.

\begin{table}
\caption{Computational environment used for model training.}
\label{tab:computational_environment}
\begin{tabular*}{\textwidth}{@{} l l @{}}
\toprule
Component & Specification \\
\midrule
GPU & NVIDIA GeForce GTX 1080 Ti \\
GPU memory & 11 GB \\
CPU & Intel Core i7-8700 @ 3.20 GHz \\
CPU cores / threads & 6 cores / 12 threads \\
RAM & 31 GiB \\
Operating system & Ubuntu 24.04.4 LTS \\
GPU driver & 580.126.09 \\
CUDA version reported by \texttt{nvidia-smi} & 13.0 \\
\bottomrule
\end{tabular*}
\end{table}

\subsection{Training dynamics}\label{subsec:training_dynamics}

\begin{figure}[htb]
    \centering
      \centering
    \subfloat[Training and validation loss curves]{\includegraphics[width=0.23\linewidth]{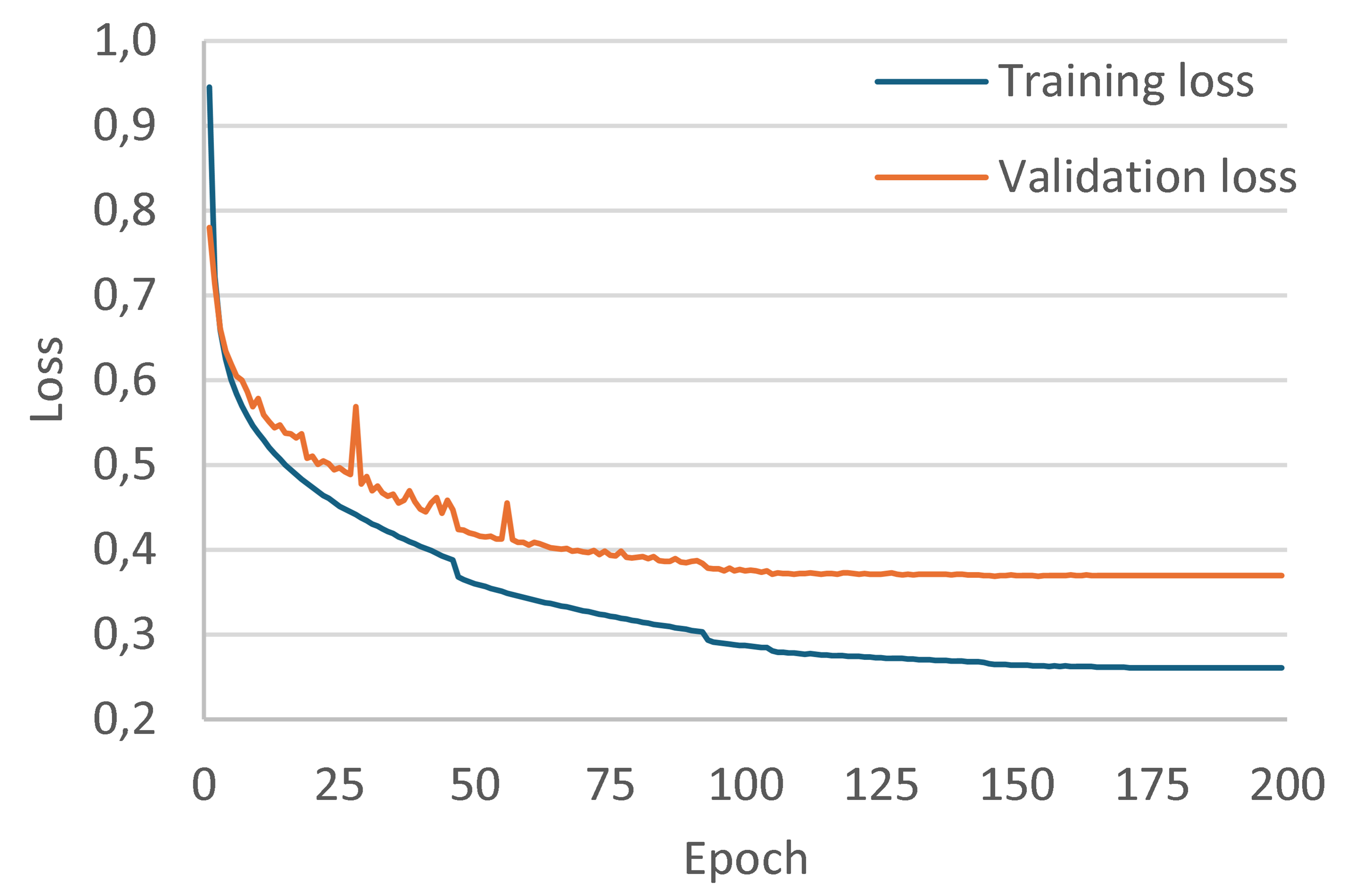}
    \label{fig:training_loss_curves}}
    \hfill
    \subfloat[Validation mIoU evolution]{\includegraphics[width=0.23\linewidth]{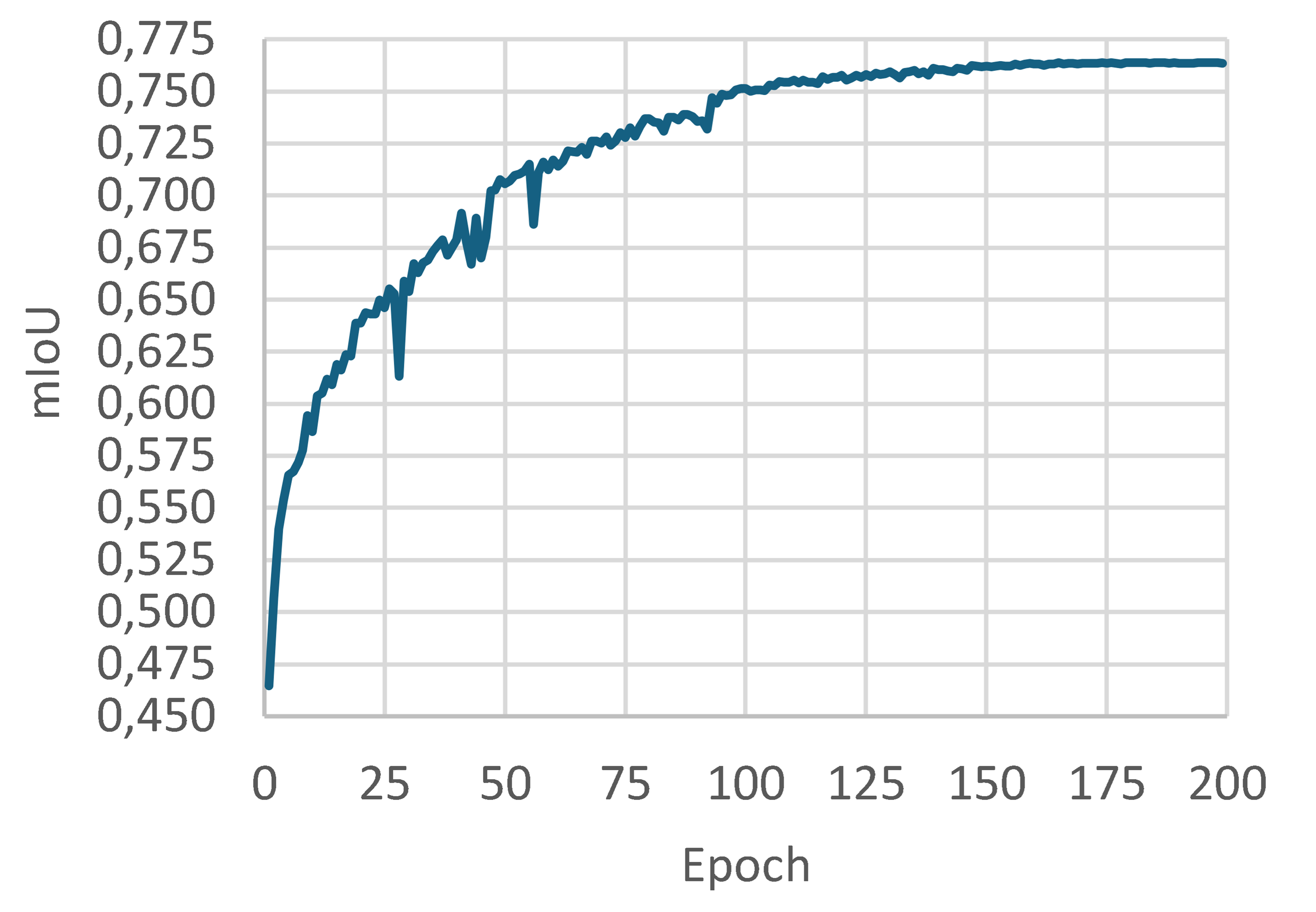}
    \label{fig:validation_miou_curve}}
    \hfill
    \subfloat[Validation pixel accuracy evolution]{\includegraphics[width=0.23\linewidth]{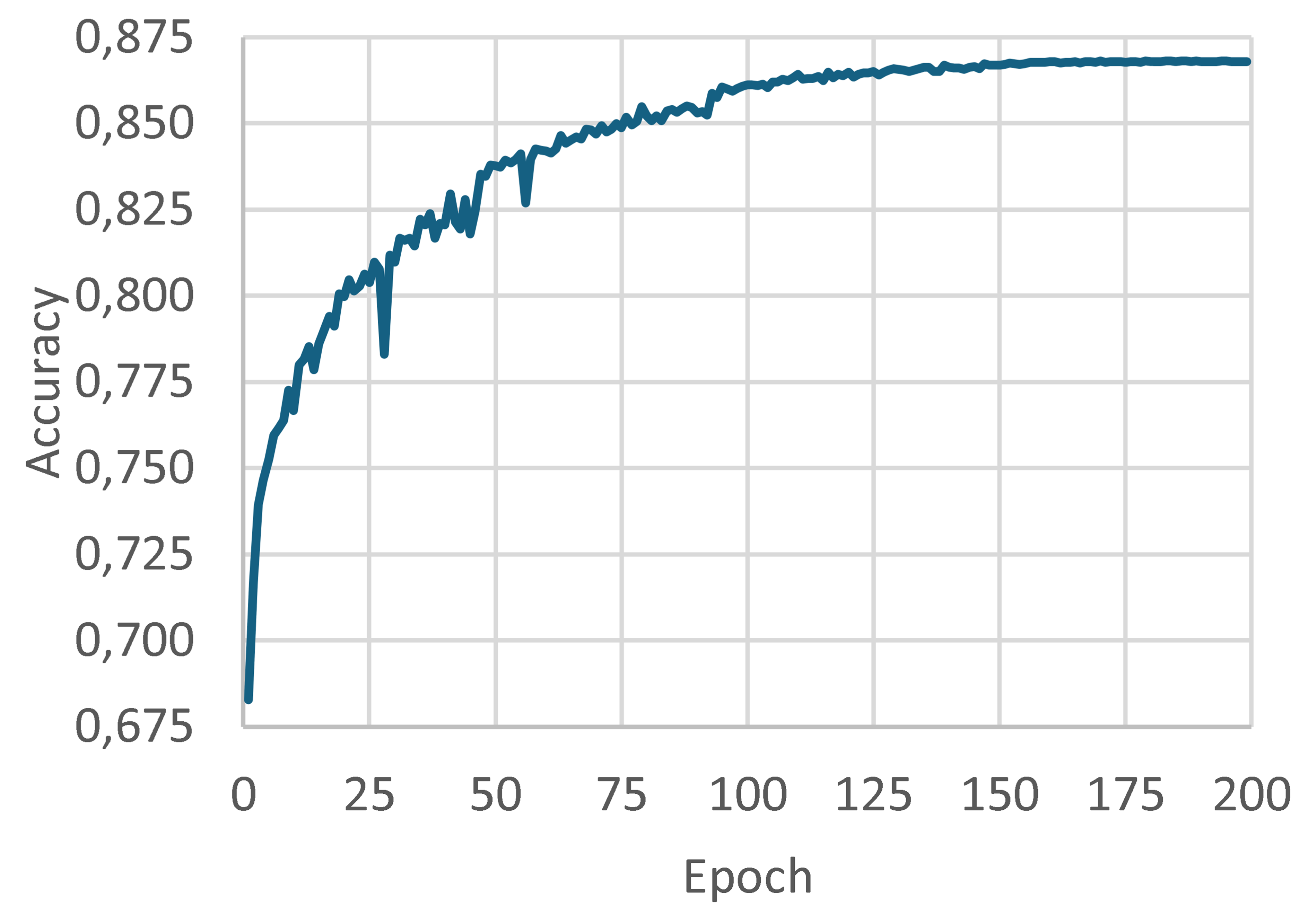}
    \label{fig:validation_pixel_accuracy_curve}}
    \hfill\subfloat[Learning rate schedule used during the training of the proposed U-Net model]{
     \includegraphics[width=0.23\linewidth]{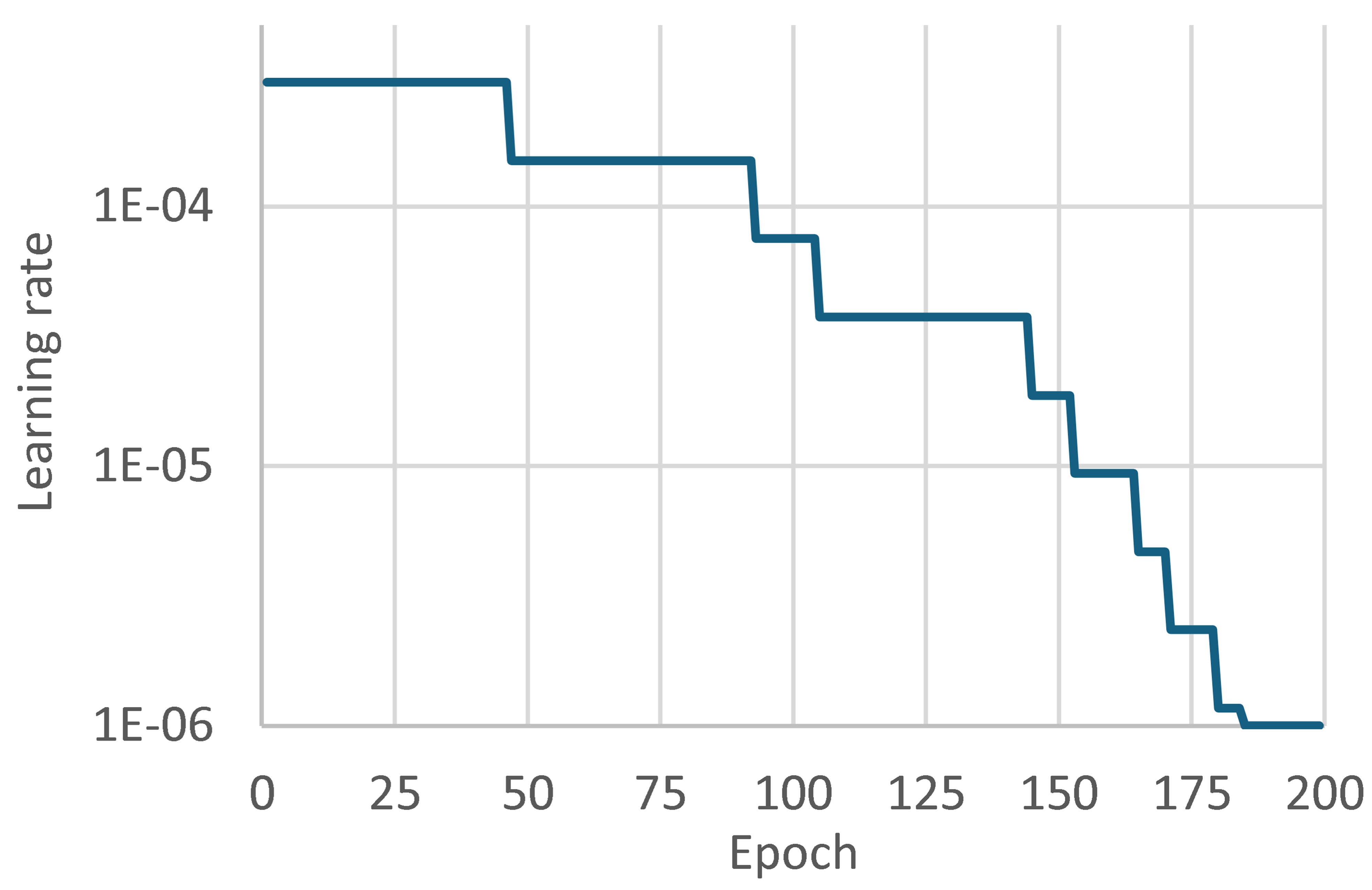}
         \label{fig:learning_rate_curve}}
    \caption{Metric evolution for the proposed U-Net model}
    
    \label{fig:training_metrics}
\end{figure}

Figure~\ref{fig:training_metrics} illustrates the evolution of the primary training and validation metrics for the proposed U-Net architecture. As shown in Figure~\ref{fig:training_loss_curves}, both training and validation losses decrease rapidly during the initial phase, indicating that the model efficiently learns meaningful representations from the image–mask pairs. Progressing into the later epochs, the validation loss gradually stabilizes into a plateau, while the training loss continues to steadily decline. The resulting gap between the two curves suggests a minor degree of overfitting, which is expected given the high complexity of the multispectral segmentation task and the heterogeneity of the dataset.

The validation mean Intersection over Union (mIoU) and pixel accuracy trends reinforce these convergence dynamics. As shown in Figure~\ref{fig:validation_miou_curve}, the validation mIoU rises consistently, reaching a peak of approximately 0.7639 at epoch 195. However, the optimal checkpoint for testing was selected at epoch 179, corresponding to the minimum validation loss. Because the mIoU values at epochs 179 and 195 were nearly identical, this choice ensures optimal generalization without sacrificing performance. Concurrently, the rapid early increase and subsequent stabilization of pixel accuracy (Figure \ref{fig:validation_pixel_accuracy_curve}) confirm progressive improvement at the pixel level, while the mIoU curve provides a more stringent evaluation of regional segmentation quality.

This optimization process was supported by a learning rate scheduler that triggered reductions when the validation loss plateaued. As illustrated in Figure~\ref{fig:learning_rate_curve}, the initial learning rate of $3 \times 10^{-4}$ was adaptively scaled down to $10^{-6}$. This strategy enabled large, efficient weight updates during early training phases and finer optimization refinements as the model approached convergence.

\subsection{Random Forest comparative baselines}\label{subsubsec:radom_forest_comparative_baselines}

In addition to the U-Net model, two Random Forest baselines were considered in order to provide a classical machine learning comparison. Random Forests are ensemble models composed of multiple decision trees, where the final prediction is obtained by aggregating the predictions of individual trees \citep{breiman2001random}. In the context of crop mapping from Sentinel-2 imagery, Random Forests are useful baselines because they can model non-linear relationships between spectral bands without relying on deep neural architectures.

The first baseline is a spectral pixel-wise Random Forest. In this setting, each pixel is treated as an independent training sample, and its input features are given by the selected Sentinel-2 spectral bands.

The target label is the corresponding class value from the rasterized segmentation mask. This baseline does not use explicit spatial context and is therefore used to evaluate how much crop discrimination can be achieved from spectral information alone.

The second baseline is a spatial-context Random Forest, which keeps the same original 10 Sentinel-2 spectral bands, but augments them with handcrafted local spatial statistics. More specifically, local mean and local standard deviation features are computed over square neighbourhoods of size $3 \times 3$ and $5 \times 5$. The local mean describes the average spectral response around a pixel, while the local standard deviation measures how much the neighbouring pixels vary around that mean. Therefore, these features provide simple information about local homogeneity, texture, and possible transitions near parcel boundaries.

To keep the model size and memory requirements manageable, spatial statistics were not computed for all spectral bands. Instead, they were computed for four informative bands: B04, B08, B11, and B12, corresponding to the red, near-infrared, and short-wave infrared regions. The resulting feature vector combines the original 10 spectral values with 16 additional spatial-context features, derived from two window sizes and two local statistics computed across the four selected bands, yielding a total of 26 features per pixel.

Both Random Forest baselines were trained on a stratified subset of pixels extracted from the training split. The sampling strategy was class-aware: for each class, a fixed target number of pixels was selected, while the number of pixels extracted from the same class inside a single patch was limited. This prevents the training table from being dominated by background or by highly frequent crop classes, and also encourages the sampled pixels to come from a large number of patches.

Bootstrap sampling was enabled during Random Forest training. This means that each decision tree was trained on a random sample of the selected training pixels, rather than on the exact same training table. In combination with $max\_samples = 0.8$ , each tree used approximately 80\% of the selected pixels, increasing diversity among trees and reducing overfitting.

During evaluation, the trained Random Forest models were applied pixel-wise to full image patches. The predicted pixels were then reshaped back into segmentation masks, and the same segmentation-oriented metrics used for U-Net were computed, including mean IoU, foreground mean IoU, pixel accuracy, and class-wise IoU. The two Random Forest baselines were designed to isolate the contribution of spectral information and handcrafted local spatial context, while U-Net evaluates the benefit of learned multi-scale spatial representations.

\subsection{Evaluation metrics}\label{subsubsec:evaluation_metrics}

Model performance was evaluated using standard semantic segmentation metrics, with mean Intersection over Union (mIoU) used as the main criterion \citep{long2015fully}. Class-wise IoU measures the overlap between the predicted and reference regions for each class, while mIoU summarizes this performance by averaging IoU values across the evaluated classes. This metric is particularly relevant for segmentation because it penalizes both missed regions and false positive predictions, providing a stricter assessment of spatial overlap than pixel accuracy.

In addition to all-class mIoU, foreground mIoU was reported for the internal comparison. This metric averages IoU only over the agricultural crop classes, excluding the background class, and therefore provides a clearer view of crop-specific segmentation performance in the presence of a large background category.

Pixel accuracy and mean class accuracy were also reported as complementary metrics. Pixel accuracy measures the overall proportion of correctly classified pixels, but it can be strongly influenced by dominant classes such as background or frequent crops. Mean class accuracy evaluates class-wise recall by averaging the accuracy of each class independently, but unlike IoU, it does not directly penalize false positive predictions. For this reason, the analysis focuses mainly on mIoU and class-wise IoU, while pixel accuracy and mean class accuracy are used as supporting indicators.

To handle varying class availability across external datasets, present-class mIoU was also reported. This metric is computed only over classes containing at least one ground-truth pixel in the evaluated subset. It is particularly relevant for external benchmarks such as DACIA5, where some target classes from the proposed taxonomy, such as barley and oats, are absent from the selected evaluation subset. Pixels assigned to the ignore value 255 in external masks were excluded from all metric computations, as they denote unlabeled or non-comparable regions rather than valid output classes.

\subsection{External evaluation datasets}\label{subsec:external_evaluation_datasets}

In addition to the internal EuroCrops-based test split, three external evaluation settings were prepared in order to assess the generalization capacity of the trained model. The first setting is an in-domain external evaluation based on unseen EuroCrops subsets from Belgium. The other two settings are cross-dataset evaluations based on DACIA5 and PASTIS, which differ more substantially from the training data in terms of geographic region, label taxonomy, temporal organization, and preprocessing assumptions.

For the Belgian EuroCrops subsets and DACIA5, the input arrays were prepared using the same reflectance range as the training data. For PASTIS, the adapted Sentinel-2 arrays were first clipped to the $[0,10000]$ range before normalization, because some observations contained negative reflectance-like values or values above the range used by the EuroCrops-derived training patches.

No dataset-specific normalization statistics were computed from the external test sets, in order to avoid adapting the model to the evaluation data. This step is particularly important for cross-dataset evaluation, because differences in band order, reflectance scaling, or preprocessing conventions may strongly affect model predictions.

\subsubsection{Unseen EuroCrops country-year subsets}\label{subsubsec:unseen_eurocrops_country_year_subsets}

To evaluate spatial generalization within the same annotation framework, two Belgian EuroCrops subsets were selected: BE\_VLG\_2021, corresponding to Flanders, and BE\_WAL\_2021, corresponding to Wallonia. These subsets were not used during training, validation, or model selection. The training dataset was constructed from Austria, France, Germany, Slovakia, and the Czech Republic; therefore, the Belgian subsets represent unseen EuroCrops regions.

The same data generation pipeline was applied to both Belgian subsets, including label harmonization, Sentinel-2 product selection, cloud and quality filtering, rasterization of parcel annotations, and extraction of $256 \times 256$ image--mask pairs. This ensured that the resulting test sets were compatible with the model input format while remaining independent from the data used for training.

Using the natural grid-based generation strategy, 7093 test patches were obtained from BE\_VLG\_2021 and 3613 test patches from BE\_WAL\_2021, resulting in a total of 10706 unseen EuroCrops test patches. These samples were used to evaluate the model under realistic class distributions specific to two Belgian agricultural regions.

Since the class distributions of these subsets may differ from both the training data and from each other, they provide useful information about the robustness of the model to regional shifts within the EuroCrops data source. In this study, the Belgian subsets are considered in-domain external tests, distinct from the cross-dataset evaluations performed on DACIA5 and PASTIS.

\subsubsection{DACIA5}\label{subsubsec:dacia5}

The DACIA5 agricultural dataset \citep{dacia5paper,dacia5dataset} comprises Sentinel-1 and Sentinel-2 imagery acquired north of Bra\c{s}ov, Romania (2020–2024), along with yearly crop masks and parcel annotations. In this study, it was used exclusively for testing as an independent benchmark to evaluate cross-dataset generalization. To maintain consistency with our training configuration, only the Sentinel-2 component was utilized. Furthermore, rather than using its original $32 \times 32$ classification patches, we considered the full 12-band GeoTIFF images and yearly reference masks to construct an external semantic segmentation benchmark, testing the U-Net model on geographically independent data with a distinct annotation source.

To align the original DACIA5 taxonomy with the reduced class system adopted in this study, we remapped the labels. The conversion to the target taxonomy is detailed in Table~\ref{tab:dacia5_label_remapping}. For labels with a reliable correspondence, the last column reports the corresponding project class ID. Labels without a reliable match were assigned the ignore value 255 and excluded from metric computation.

\begin{table}[]
\caption{DACIA5 label remapping used for the external semantic segmentation evaluation. }
\label{tab:dacia5_label_remapping}
\footnotesize

\begin{tabular*}{\textwidth}{@{} p{2.7cm} p{10cm} r @{}}
\toprule
Target class & DACIA5 labels & Project label ID \\
\midrule
Maize & Corn and corn silage & 1 \\
Rapeseed & Rapeseed & 2 \\
Wheat & Winter wheat and spring wheat & 3 \\
Sunflower & Sunflower & 6 \\
Sugar beet & Sugar beet & 7 \\
Soybean & Soybean & 8 \\
Potato & Potato-related classes & 9 \\
Pasture & Temporal grassland, permanent grassland, and alfalfa-related classes & 10 \\
Ignored & Peas and labels without a reliable correspondence in the selected target taxonomy & 255 (ignored) \\
\bottomrule
\end{tabular*}
\end{table}

The ignore value was used only during external evaluation and was not part of the model output space. The U-Net model was trained to predict the 11 project classes, while pixels marked as 255 in the adapted DACIA5 masks were excluded before computing the evaluation metrics. This choice was necessary because some DACIA5 labels, such as peas, correspond to real agricultural crops but do not have a reliable correspondence in the proposed target taxonomy. Mapping such pixels to background would incorrectly treat them as non-crop or non-target areas.

The DACIA5 Sentinel-2 images are distributed as 12-band GeoTIFF files at 10 m spatial resolution \citep{dacia5dataset}. Before patch extraction, the band order was inspected and validated against a reference Sentinel-2 SAFE product corresponding to the same acquisition date and region. Based on this verification, the ten spectral bands used by the proposed model were selected from the DACIA5 stack, namely B02, B03, B04, B08, B05, B06, B07, B8A, B11, and B12.

To construct the external test benchmark, the yearly DACIA5 images and masks were divided into fixed patches of size $256 \times 256$ pixels, consistent with the input size used during training. Patch extraction was performed using a sliding-window strategy over each annual image, including border-aligned windows to ensure full spatial coverage. Patches containing only ignored labels or only background were discarded, and only samples containing at least 1000 valid labeled crop pixels were retained.

Using this procedure, a total of 1172 patches were generated from 172 Sentinel-2 scenes covering the period 2020--2024. In total, 1376 candidate windows were considered, of which 1172 were retained and 204 were discarded after filtering. For the experiments reported in this work, evaluation was restricted to the April--September seasonal window, resulting in 778 evaluated patches. These values correspond to the external benchmark constructed in this work and not to the original patch-level splits provided by DACIA5.

An important aspect of the DACIA5 evaluation concerns the interpretation of background pixels. The reference masks originate from specific parcels managed by the National Institute of Research and Development for Potato and Sugar Beet (Bra\c{s}ov, Romania), rather than from exhaustive annotations of the entire scene. Consequently, a background label does not necessarily indicate a non-crop pixel; it may simply correspond to an agricultural field missing from the reference data.

For this reason, the main DACIA5 evaluation reported in this work focuses on labeled crop pixels only. Background and ignored pixels were excluded from the main metric computation in order to evaluate transfer performance on crop classes with reliable reference labels and to avoid penalizing predictions in areas where no crop annotation is available.

\subsubsection{PASTIS}\label{subsubsec:pastis}

Alongside DACIA5, the PASTIS dataset \citep{garnot2021pastis,pastis_dataset} serves as a second cross-dataset benchmark to evaluate generalization under an alternative data organization and annotation protocol. Comprising 2,433 annotated sequences across metropolitan France, PASTIS provides pixel-level semantic and instance annotations for agricultural parcels. While our EuroCrops-based dataset is structured for static multispectral segmentation, PASTIS functions as a temporally structured Sentinel-2 benchmark. It contains multiple observations per patch throughout the agricultural season, making it especially suited for evaluating crop mapping models under realistic temporal and atmospheric variability.

In the context of this work, the PASTIS dataset was adapted to a fixed-input multispectral setting rather than being evaluated under its original benchmark protocol. This adaptation was necessary because the original PASTIS class taxonomy does not directly match the reduced target taxonomy adopted in this study, and its temporal organization differs substantially from the static, patch-based inputs used to train our U-Net model. Consequently, integrating PASTIS required an additional preprocessing pipeline encompassing class remapping, temporal selection, band reordering, and mask adaptation. Specifically, the original semantic labels were mapped to our target taxonomy whenever a meaningful correspondence existed; classes lacking a reliable alignment were assigned an ignore label and excluded from metric computations. The complete taxonomic mapping is detailed in Table \ref{tab:pastis_label_remapping}.

\begin{table}[]
\caption{PASTIS label remapping used for the external single-date semantic segmentation evaluation. Labels without a reliable correspondence in the proposed target taxonomy were assigned to the ignore value and excluded from metric computation.}
\label{tab:pastis_label_remapping}
\footnotesize
\begin{tabular*}{\textwidth}{@{\extracolsep{\fill}} c p{4.7cm} p{4.0cm} r @{}}
\toprule
PASTIS ID & Original PASTIS label & Mapped project class & Project ID / treatment \\
\midrule
0  & Background & Background & 0 \\
1  & Meadow & Pasture & 10 \\
2  & Soft winter wheat & Wheat & 3 \\
3  & Corn & Maize & 1 \\
4  & Winter barley & Barley & 4 \\
5  & Winter rapeseed & Rapeseed & 2 \\
6  & Spring barley & Barley & 4 \\
7  & Sunflower & Sunflower & 6 \\
8  & Grapevine & Ignored & 255 \\
9  & Beet & Sugar beet & 7 \\
10 & Winter triticale & Wheat & 3 \\
11 & Winter durum wheat & Wheat & 3 \\
12 & Fruits, vegetables, and flowers & Ignored & 255 \\
13 & Potatoes & Potato & 9 \\
14 & Leguminous fodder & Pasture / forage-related class & 10 \\
15 & Soybeans & Soybean & 8 \\
16 & Orchard & Ignored & 255 \\
17 & Mixed cereal & Ignored & 255 \\
18 & Sorghum & Ignored & 255 \\
19 & Void label & Ignored & 255 \\
\bottomrule
\end{tabular*}
\end{table}

The ignore value (255) was used only during external evaluation and was omitted from the U-Net model's 11-class output space. This choice prevents unfair evaluation penalties; several PASTIS classes lacking a direct target correspondence (e.g., grapevine, orchard, sorghum, fruits/vegetables) still represent agricultural surfaces rather than true background, making it inaccurate to map them as such.

To match the U-Net model trained on individual Sentinel-2 observations, PASTIS was adapted to a single-date setting. Observations from the April–September seasonal interval were converted to the standard 10-band input representation, reordered to match the EuroCrops-based training format, and remapped to the target taxonomy. This adaptation yielded 1,955 single-date evaluation samples.

During adaptation, the radiometric range of the PASTIS arrays was inspected. Because some observations contained reflectance values outside the approximate $[0, 10000]$ range used to train the U-Net model, the PASTIS images were clipped to this interval before normalization. This preprocessing step ensures consistency with the training representation without altering the reference masks.

Furthermore, while the training patches measure $256 \times 256$ pixels, the original PASTIS patches are $128 \times 128$ pixels. Although the fully convolutional U-Net handles variable input sizes during inference, this spatial difference introduces an additional source of mismatch between the training and evaluation settings.

This setup departs from the original PASTIS temporal benchmark protocol and instead serves as a strict cross-dataset and cross-protocol transfer evaluation, where a model trained on single-date EuroCrops-derived samples is tested on data originally designed for multi-temporal segmentation. This adapted use of PASTIS provides a complementary external evaluation scenario, distinct from both the EuroCrops-based dataset and the DACIA5 benchmark, and is expected to be substantially more challenging, as discriminating crops from a single acquisition foregoes the phenological information that time series naturally encode.

\section{Results and discussion}\label{sec:results_and_discutions}

The model was evaluated using five test datasets organized into three complementary evaluation categories, as summarized in Table~\ref{tab:evaluation_settings}. The first category corresponds to the internal evaluation, performed on the EuroCrops-based test split generated by the proposed pipeline. The second category corresponds to in-domain external evaluation, using two unseen Belgian EuroCrops subsets, BE\_VLG\_2021 and BE\_WAL\_2021. These subsets follow the same annotation source and preprocessing workflow, but correspond to regions not used during training, validation, or model selection. The third category corresponds to cross-dataset external evaluation, using DACIA5 and PASTIS, which differ more substantially from the training data in terms of geographic region, label taxonomy, temporal organization, and data preparation assumptions.

\begin{table}[]
\caption{Overview of the evaluation settings used in this study.}
\label{tab:evaluation_settings}
\footnotesize
\begin{tabular*}{\textwidth}{@{\extracolsep{\fill}} p{3.2cm} p{3.2cm} c p{7.0cm} @{}}
\toprule
Evaluation category & Dataset & Patches & Purpose \\
\midrule
Internal evaluation 
& EuroCrops test split 
& 6652 
& Internal performance on the generated dataset. \\

In-domain external 
& BE\_VLG\_2021 
& 7093 
& Spatial generalization to an unseen EuroCrops region. \\

In-domain external 
& BE\_WAL\_2021 
& 3613 
& Spatial generalization to another unseen EuroCrops region. \\

Cross-dataset external 
& DACIA5 
& 778 
& Transfer to an independent Romanian dataset using the April--September evaluation subset. \\

Cross-dataset external 
& PASTIS 
& 1955 
& Transfer to a temporally structured benchmark adapted to a single-date April--September evaluation setting. \\
\bottomrule
\end{tabular*}
\end{table}

\subsection{Internal evaluation of the proposed class-weighted U-Net model}\label{subsec:internal_evaluation_of_the_proposed_class_weighted_unet_model}

The proposed class-weighted U-Net model was evaluated on the internal EuroCrops-based test split generated from the final manifest. The final manifest contains 67,337 selected patches, divided into 53,888 training patches, 6,797 validation patches, and 6,652 test patches. The test split was not used during training or model selection.

The proposed class-weighted U-Net model achieved a test mIoU of 0.7665, a pixel accuracy of 0.8693, and a mean class accuracy of 0.9072 on the internal EuroCrops test split. These results demonstrate that the generated dataset supports robust crop segmentation and that the model segments the target classes with good accuracy. Nevertheless, the task remains challenging due to class imbalance, spectral similarities between certain crops, and the heterogeneous nature of parcel-level annotations.

The discrepancy between pixel accuracy and mean class accuracy underscores the necessity of balanced evaluation. While pixel accuracy is heavily influenced by dominant classes such as background and wheat, mean class accuracy evaluates performance more equitably across categories. Consequently, we assess performance primarily through mIoU and class-wise IoU, treating pixel and mean class accuracy as complementary indicators. This class-wise analysis is presented in Table~\ref{tab:unet_class_iou}.

\begin{table}[]
\caption{Class-wise performance of the proposed U-Net model on the EuroCrops test split.}
\label{tab:unet_class_iou}
\footnotesize
\begin{tabular*}{\textwidth}{@{}@{\extracolsep{\fill}} l l l l @{}}
\toprule
Class & IoU & Class accuracy & Support pixels \\
\midrule
Background & 0.7352 & 0.7647 & 136,879,857 \\
Maize & 0.8334 & 0.9404 & 47,844,217 \\
Rapeseed & 0.8161 & 0.9317 & 26,035,062 \\
Wheat & 0.8075 & 0.9225 & 102,534,883 \\
Barley & 0.7045 & 0.8457 & 37,137,182 \\
Oats & 0.6583 & 0.9042 & 5,832,341 \\
Sunflower & 0.7887 & 0.9476 & 9,282,741 \\
Sugar beet & 0.8553 & 0.9507 & 18,081,916 \\
Soybean & 0.7700 & 0.9352 & 10,796,959 \\
Potato & 0.7599 & 0.9364 & 9,469,228 \\
Pasture & 0.7021 & 0.8990 & 32,051,086 \\
\bottomrule
\end{tabular*}
\end{table}

The highest IoU values were obtained for sugar beet, maize, rapeseed, and wheat, showing that the proposed model is able to segment several major crop categories with good spatial overlap. Sunflower, soybean, and potato also achieved relatively strong results, despite having lower pixel-level support than dominant classes such as background and wheat. Class accuracy is generally higher than IoU because it measures the proportion of correctly recovered pixels within each ground-truth class, while IoU also penalizes false positive predictions. Therefore, IoU remains the main class-wise metric, whereas class accuracy is used as complementary information about class-level recall.

Lower IoU values were obtained for oats, pasture, and barley. The result for oats is consistent with its reduced support in the test set and its spectral similarity to other cereal crops. Barley is also affected by similarity to wheat and other cereals, while pasture is a more heterogeneous class that may include visually diverse grassland and forage-related areas. These results confirm that the main remaining challenges are related to cereal crop separation, heterogeneous vegetation classes, and the limited representation of some minority crops.

\subsubsection*{Confusion analysis}\label{subsec:confusion_analysis}

\begin{figure}
    \centering
    \includegraphics[width=0.7\linewidth]{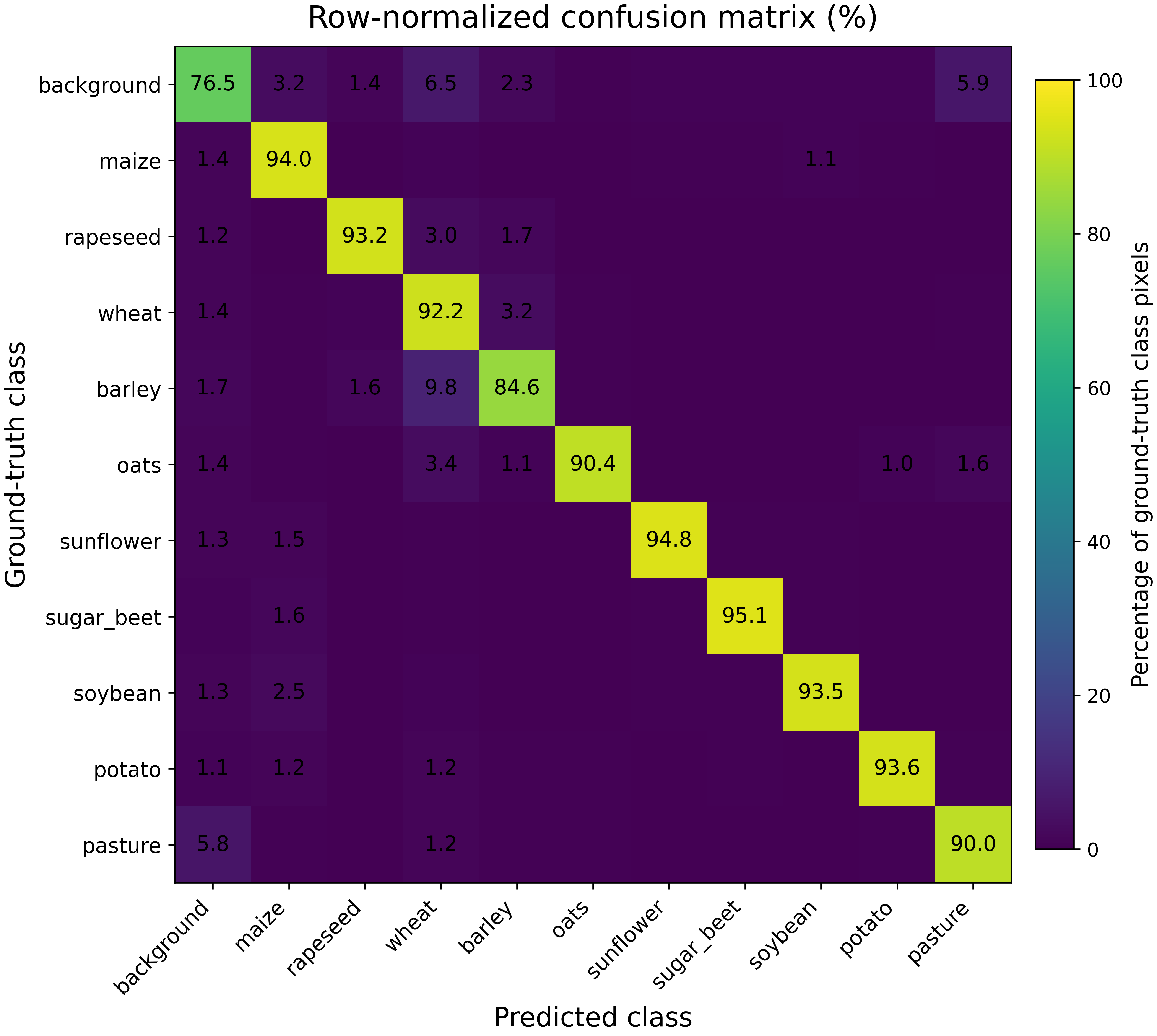}
    \caption{Row-normalized confusion matrix for the proposed U-Net model on the internal EuroCrops test split. Values are expressed as percentages, with rows corresponding to ground-truth classes and columns corresponding to predicted classes.}
    \label{fig:confusion_matrix_percent}
\end{figure}

To better understand the class-wise behaviour of the model, a row-normalized confusion matrix was computed and is illustrated in Figure \ref{fig:confusion_matrix_percent}. Each row corresponds to a ground-truth class, and each entry indicates the percentage of pixels from that class assigned to each predicted category. This normalization makes the matrix more interpretable under class imbalance, as each class is evaluated independently of its pixel count.

The confusion matrix supports the class-wise results reported in Table~\ref{tab:unet_class_iou}. Several crop categories show strong diagonal values, indicating that most pixels belonging to these classes are correctly recovered. The strongest class accuracies are obtained for sugar beet, sunflower, maize, potato, soybean, rapeseed, and wheat. These results suggest that the model learns meaningful spectral and spatial patterns for both dominant crops and several less frequent classes.

At the same time, the confusion matrix highlights the main remaining sources of error. Background has a lower class accuracy than most crop classes, indicating that some crop pixels and non-target regions remain difficult to separate. Barley and oats are also more challenging, which is consistent with their spectral similarity to other cereal crops, especially wheat. Pasture presents additional ambiguity because it represents a heterogeneous vegetation category rather than a single well-defined crop type.

Overall, the confusion analysis confirms that the proposed model performs well on several major crop categories, while the most persistent errors are related to cereal crop separation, background ambiguity, and heterogeneous vegetation classes such as pasture.

\subsubsection*{Qualitative segmentation examples}\label{subsec:qualitative_segmentation_examples}

\begin{figure}[htbp]
    \centering
    \includegraphics[width=0.95\textwidth]{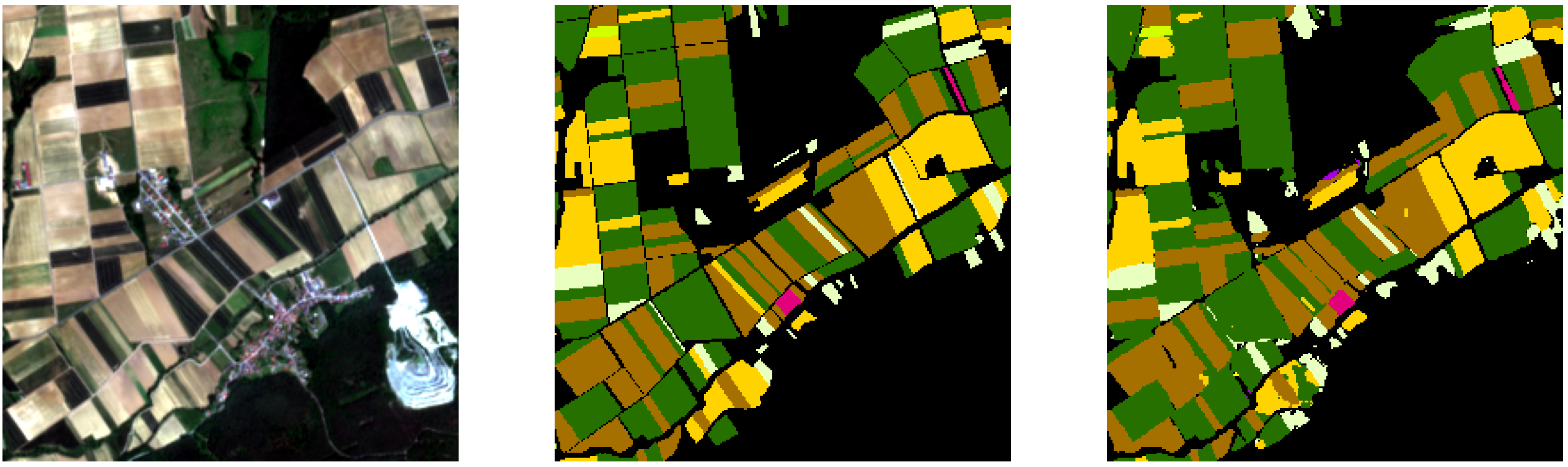}

    \vspace{0.25cm}

    \includegraphics[width=0.95\textwidth]{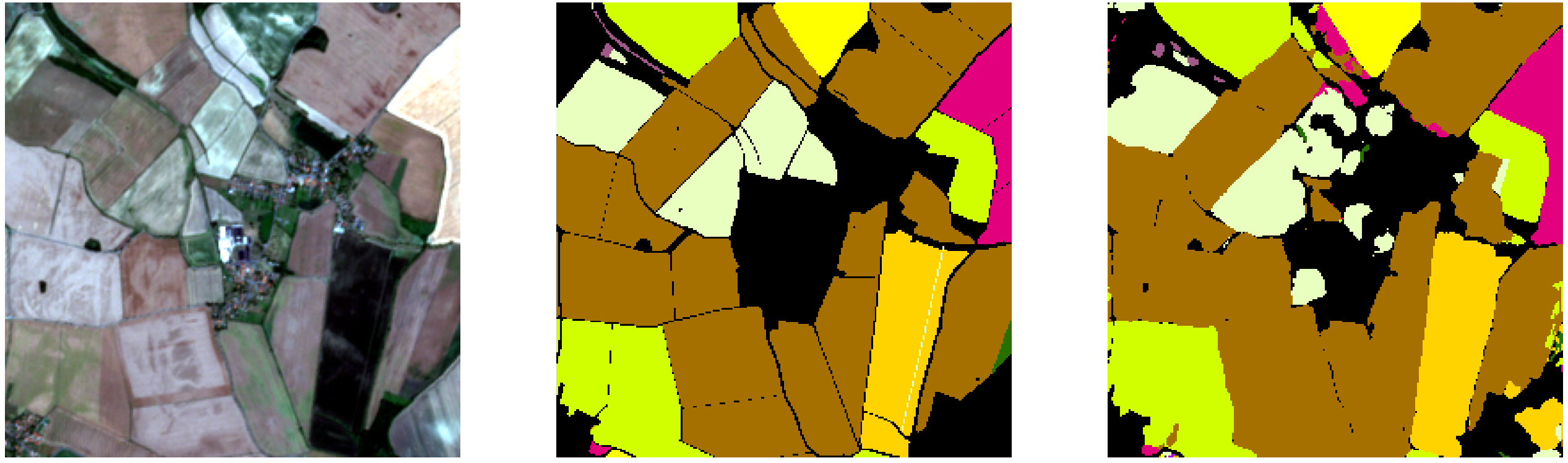}

    \caption{Qualitative U-Net segmentation examples on the internal EuroCrops test split. Each row shows, from left to right, the Sentinel-2 RGB visualization, the ground-truth segmentation mask, and the U-Net prediction.}
    \label{fig:unet_qualitative_predictions}
\end{figure}

To complement the internal quantitative evaluation, qualitative predictions were inspected on samples from the internal EuroCrops test split. Figure~\ref{fig:unet_qualitative_predictions} shows representative examples comparing the Sentinel-2 RGB visualization, the ground-truth segmentation mask, and the U-Net prediction. The same class-colour encoding as in Figure~\ref{fig:rgb_mask} is used.

The qualitative examples in Figure~\ref{fig:unet_qualitative_predictions} show that the proposed U-Net generally produces spatially coherent segmentation masks. Large agricultural parcels are often recovered as consistent regions, and the predicted masks preserve much of the parcel-level structure visible in the ground-truth annotations.

Most errors appear near parcel boundaries, in narrow field structures, or in areas where neighbouring crop classes have similar visual and spectral characteristics. This behaviour is expected in crop segmentation from Sentinel-2 imagery, especially when several agricultural classes are spatially close and spectrally similar.

The qualitative results are consistent with the quantitative analysis. Classes with stronger IoU values, such as sugar beet, maize, rapeseed, and wheat, generally appear more stable in the predicted masks, while more difficult categories such as barley, oats, and pasture remain affected by confusion and boundary errors.

Overall, the qualitative inspection confirms that the U-Net model benefits from learned spatial context and produces more coherent segmentation outputs than would be expected from purely pixel-wise classification. These examples complement the numerical metrics by illustrating how the model behaves at parcel level and where the main visual errors occur.

\subsection{Comparison with Random Forest baselines}\label{subsec:comparison_with_random_forest_baselines}

To evaluate the contribution of learned spatial representations, the U-Net model was compared with two Random Forest baselines. The first Random Forest uses only the spectral values of each pixel, while the second one augments the spectral representation with local mean and standard deviation features computed over $3 \times 3$ and $5 \times 5$ neighbourhoods.

This comparison separates three levels of information used for crop segmentation. The spectral Random Forest evaluates the discriminative power of individual Sentinel-2 pixel spectra. The spatial Random Forest evaluates whether handcrafted local context improves over the purely spectral representation. Finally, U-Net evaluates the benefit of learning multi-scale spatial features directly from image patches.

\begin{table}[]
\caption{Comparison between U-Net and Random Forest baselines on the internal EuroCrops test split.}
\label{tab:rf_unet_internal_comparison}
\footnotesize
\begin{tabular*}{\textwidth}{@{\extracolsep{\fill}} l p{4.0cm} p{3.0cm} c c @{}}
\toprule
Model & Feature type & Spatial context & All-class mIoU & Pixel accuracy \\
\midrule
Random Forest & Spectral bands & No & 0.2521 & 0.3975 \\
Random Forest & Spectral + local statistics & Handcrafted & 0.2597 & 0.4027 \\
U-Net & Multispectral patch & Learned & 0.7665 & 0.8693 \\
\bottomrule
\end{tabular*}
\end{table}

As detailed in Table \ref{tab:rf_unet_internal_comparison}, the spectral Random Forest baseline achieved an all-class mIoU of 0.2521, which improved only marginally to 0.2597 when incorporating local neighbourhood statistics via the spatial-context variant. This limited gain indicates that handcrafted features, such as local mean and standard deviation, are insufficient to capture the intricate spatial structure of agricultural parcels. Consequently, both Random Forest baselines perform substantially below the U-Net model (0.7665 mIoU), underscoring the advantage of a fully convolutional architecture that processes complete image patches to learn robust, multi-scale spatial representations directly from the data.

At the class level, the spectral Random Forest performed best on sugar beet (0.4812), maize (0.4696), wheat (0.4520), and background (0.4317), while severely struggling with barley (0.2501), soybean (0.2050), sunflower (0.2385), potato (0.1574), and oats (0.0820). These weak class-wise results—particularly for cereals like oats and barley—confirm that independent pixel-level spectral values are insufficient for robust crop segmentation, especially given the high spectral overlap and spatial heterogeneity of agricultural fields. Because this baseline predicts each pixel independently, its output is inherently fragmented and lacks spatial coherence. In contrast, the proposed U-Net model achieves a significantly superior mIoU by processing entire image patches, demonstrating that the segmentation task relies fundamentally on learning multi-scale spatial structures and parcel-level continuity rather than raw spectral separability alone.

\subsection{External generalization results}\label{subsec:external_generalization_results}

In addition to the internal EuroCrops test split, the trained models were evaluated on external datasets in order to assess their generalization capacity. The Belgian EuroCrops subsets were used to evaluate spatial generalization within the same annotation framework, while DACIA5 and PASTIS were considered as cross-dataset benchmarks with different geographic, taxonomic, and preprocessing characteristics.

The Belgian EuroCrops subsets provide an in-domain but regionally independent test, while DACIA5 and PASTIS provide stricter cross-dataset evaluations. Therefore, the internal and external results are interpreted as complementary rather than interchangeable evaluation settings.

For the external Belgian subsets, two complementary evaluation protocols were used. The first protocol evaluates all valid pixels, including the background class, and therefore reflects the complete segmentation task. The second protocol ignores background pixels and evaluates only crop pixels. This crop-pixel evaluation is useful because it separates crop-class discrimination from the additional difficulty of background--crop separation, which can vary substantially between regions.

\begin{table}[]
\caption{External evaluation on unseen Belgian EuroCrops subsets.}
\label{tab:belgian_external_results}
\footnotesize
\begin{tabular*}{\textwidth}{@{\extracolsep{\fill}} l c c c c c @{}}
\toprule
Dataset & Patches & All-pixel mIoU & All-pixel acc. & Crop-pixel mIoU present & Crop-pixel acc. \\
\midrule
BE\_VLG\_2021 & 7093 & 0.3673 & 0.7507 & 0.4427 & 0.8194 \\
BE\_WAL\_2021 & 3613 & 0.3836 & 0.7473 & 0.4944 & 0.7902 \\
\bottomrule
\end{tabular*}
\end{table}

The results show a clear performance decrease compared with the internal EuroCrops test split, which is expected because the Belgian subsets correspond to unseen regions with different crop distributions. However, the crop-pixel evaluation shows that the model retains a meaningful capacity to distinguish agricultural classes when background pixels are excluded. The crop-pixel mIoU is higher than the all-pixel mIoU on both BE\_VLG\_2021 and BE\_WAL\_2021, reaching 0.4427 and 0.4944, respectively. This suggests that part of the external performance drop is related to background--crop separation and regional distribution shifts, not only to crop-class confusion.

\begin{figure}[htbp]
    \centering
    \includegraphics[width=0.95\textwidth]{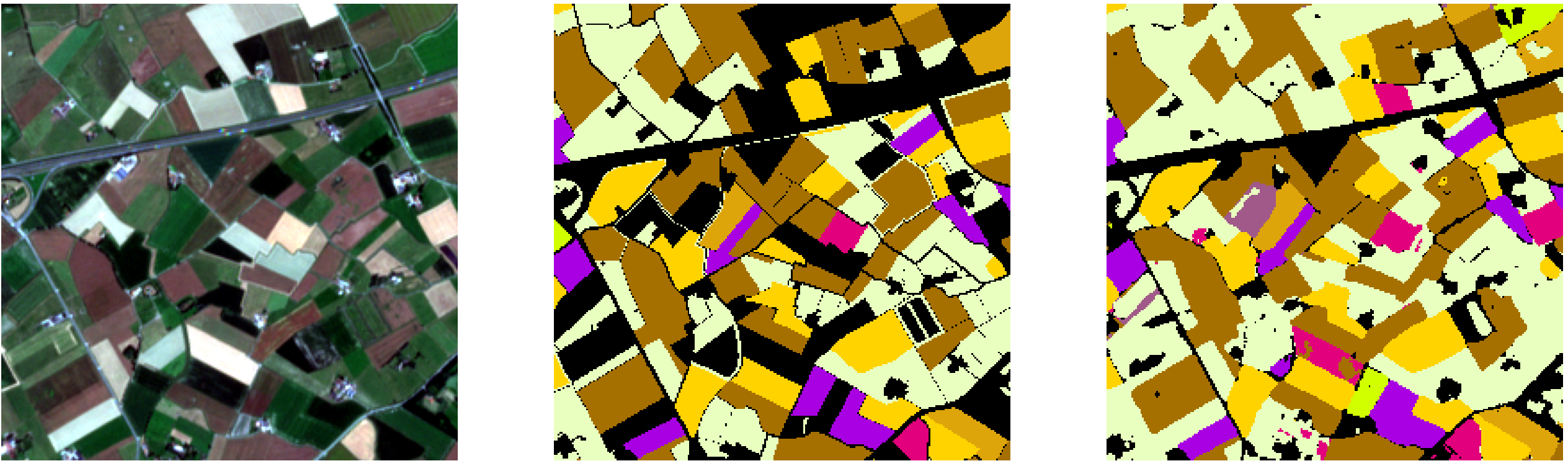}

    \vspace{0.25cm}

    \includegraphics[width=0.95\textwidth]{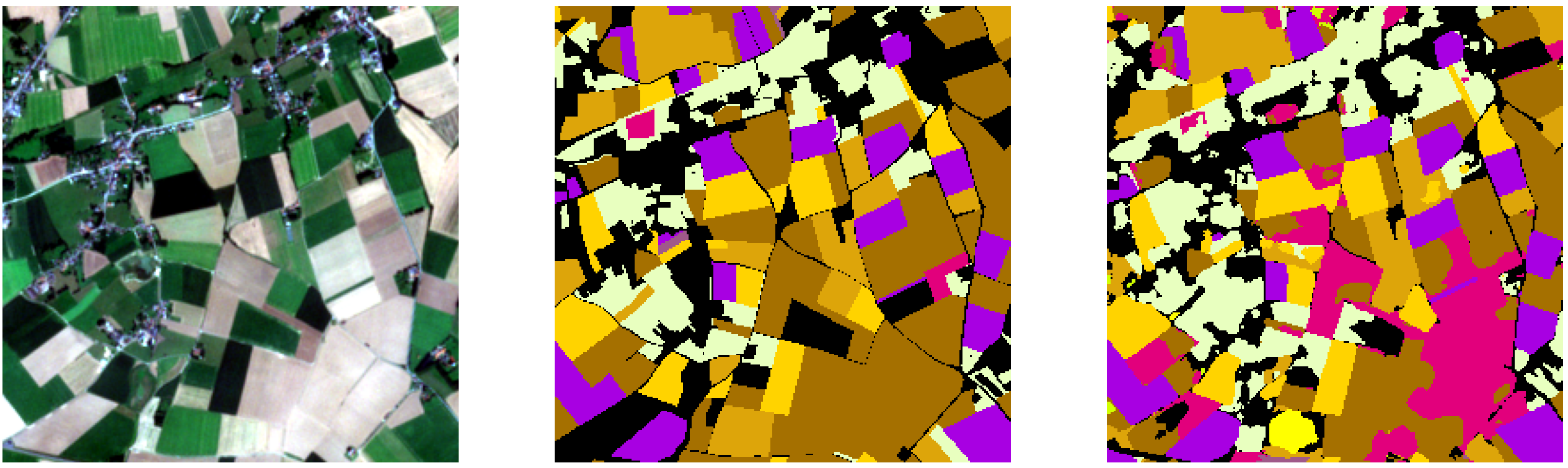}

    \caption{Qualitative prediction examples on the unseen Belgian EuroCrops subsets. Each row shows, from left to right, the Sentinel-2 RGB visualization, the ground-truth segmentation mask, and the U-Net prediction. The first example corresponds to BE\_VLG\_2021, while the second corresponds to BE\_WAL\_2021.}
    \label{fig:belgium_qualitative_predictions}
\end{figure}

The qualitative examples in Figure~\ref{fig:belgium_qualitative_predictions} support the quantitative results obtained on the unseen Belgian EuroCrops subsets. The model preserves part of the parcel-level spatial structure and produces coherent crop regions in several areas, indicating partial transfer of the learned representations to regions not used during training. At the same time, some class confusion and boundary errors remain visible, especially in areas with small parcels, neighboring fields with similar appearance, or heterogeneous background regions.

DACIA5 represents a challenging cross-dataset evaluation setting, as it differs from the training data in terms of geography, annotation source, class taxonomy, and dataset construction. Furthermore, the interpretation of background pixels differs from the internally generated EuroCrops-based dataset: in DACIA5, the reference masks are tied to institute-managed parcels, while the Sentinel-2 scenes may contain additional agricultural fields outside these annotations. Consequently, background pixels cannot be treated as reliable negative samples for evaluating crop segmentation transfer.

For this reason, two complementary evaluations were conducted on DACIA5. The first includes all valid pixels and serves as a diagnostic evaluation, while the second excludes background pixels and focuses exclusively on labeled crop pixels. The crop-pixel protocol is considered the primary DACIA5 evaluation in this work, as it assesses crop-class transfer while avoiding penalization in unlabeled regions. Since barley and oats are absent from the DACIA5 evaluation subset, present-class mIoU is the most informative metric for this benchmark.

PASTIS constitutes a second cross-dataset evaluation setting, differing from the proposed dataset in an additional respect: its original formulation is based on Sentinel-2 time series. To ensure compatibility with the trained U-Net model, PASTIS was adapted to a single-date input format, and the resulting evaluation should therefore be interpreted as a strict cross-dataset and cross-protocol transfer test. The evaluation was performed on an April–September single-date subset of 1955 samples, with background and ignored pixels excluded from the main metric computation. As with DACIA5, present-class mIoU is reported due to the incomplete class coverage of the evaluated subset.

The results for both external benchmarks are presented comparatively in Table \ref{tab:dacia5_external_results}.

\begin{table}[]
\caption{External evaluation on the DACIA5 and PASTIS April-September subsets.}
\label{tab:dacia5_external_results}
\footnotesize
\begin{tabular*}{\textwidth}{@{\extracolsep{\fill}} l c c c c @{}}
\toprule
Evaluation protocol & Patches & mIoU all project classes & mIoU present classes & Pixel acc. \\
\midrule
DACIA5 - All valid pixels & 778 & 0.1505 & 0.1839 & 0.5325 \\
DACIA5 - Labeled crop pixels only & 778 & 0.2383 & 0.3277 & 0.5509 \\
PASTIS - Labeled crop pixels only & 1955 & 0.1189 & 0.1453 & 0.3656 \\
\bottomrule
\end{tabular*}
\end{table}

On the DACIA5 benchmark, the all-valid-pixel evaluation reveals a substantial performance drop compared to the internal test split, confirming the difficulty of direct cross-dataset transfer. However, this protocol is heavily skewed by the ambiguity of DACIA5 background pixels; restricting the evaluation to labeled crop pixels increases the all-class mIoU from 0.1505 to 0.2383, and the present-class mIoU from 0.1839 to 0.3277. This shifts some of the performance gap to background handling and absent classes, leaving the remaining drop to genuine domain shift. PASTIS proves to be an even more challenging external setting, yielding a present-class mIoU of 0.1453 and a pixel accuracy of 0.3656 after radiometric clipping. This lower performance stems from a severe, multi-level domain and protocol mismatch: the U-Net was trained on static, $256 \times 256$ EuroCrops patches, whereas PASTIS features a distinct annotation protocol, an alternative taxonomy, smaller $128 \times 128$ patches, and an inherently multi-temporal design. Because the single-date model cannot exploit seasonal phenological evolution, these low scores reflect a strict, highly constrained cross-protocol stress test rather than a failure on the benchmark itself.

A class-wise analysis on both DACIA5 and PASTIS crop-pixel evaluation is reported in Table~\ref{tab:combined_external_evaluation}. Background is ignored in this protocol, while barley and oats are absent from the evaluated subset. Therefore, these classes are not considered when interpreting the present-class mIoU.

\begin{table*}[htbp]
\centering
\caption{Class-wise evaluation comparison on the DACIA5 and PASTIS benchmarks (April--September). Values reflect crop-pixel evaluations with background and ignored pixels excluded from metric computations.}
\label{tab:combined_external_evaluation}
\small
\begin{tabular*}{\textwidth}{@{\extracolsep{\fill}} p{1.7cm} c c c c r r @{}}
\toprule
 & \multicolumn{2}{c}{DACIA5 (Romania)} & \multicolumn{2}{c}{PASTIS (France)} & \multicolumn{2}{c}{Support Pixels} \\
\cmidrule(r){2-3} \cmidrule(lr){4-5} \cmidrule(l){6-7}
Class & IoU & Class Acc. & IoU & Class Acc. & DACIA5 & PASTIS \\
\midrule
Maize       & 0.5883 & 0.7902 & 0.4557 & 0.6891 & 2,266,208 & 3,423,989 \\
Rapeseed    & 0.4341 & 0.4855 & 0.0911 & 0.1163 & 543,364   & 699,069   \\
Wheat       & 0.5619 & 0.6152 & 0.2967 & 0.4465 & 3,344,695 & 3,329,186 \\
Barley      & --     & --     & 0.0679 & 0.1072 & --        & 1,015,725 \\
Sunflower   & 0.0240 & 0.3134 & 0.0063 & 0.0066 & 24,024    & 358,357   \\
Sugar beet  & 0.3340 & 0.3474 & 0.1132 & 0.1334 & 1,806,784 & 321,262   \\
Soybean     & 0.2482 & 0.3544 & 0.0117 & 0.0138 & 461,432   & 415,684   \\
Potato      & 0.0942 & 0.0982 & 0.0176 & 0.0325 & 631,207   & 98,177    \\
Pasture     & 0.3370 & 0.5997 & 0.2473 & 0.2908 & 574,014   & 7,456,450 \\
\bottomrule
\end{tabular*}
\end{table*}The best DACIA5 results are obtained for maize and wheat, followed by rapeseed, pasture, and sugar beet. These classes have stronger support in the evaluated subset and also correspond to major agricultural categories present in both the training taxonomy and the DACIA5 benchmark. In contrast, sunflower and potato obtain low IoU values. Sunflower has very limited support in the DACIA5 subset, while potato appears difficult to transfer across datasets, probably due to differences in regional crop appearance, annotation protocol, and acquisition conditions. Overall, the DACIA5 evaluation should be interpreted as a strict cross-dataset transfer test rather than as an internal validation of the proposed pipeline.

\begin{figure}[htbp]
    \centering
    \includegraphics[width=0.95\textwidth]{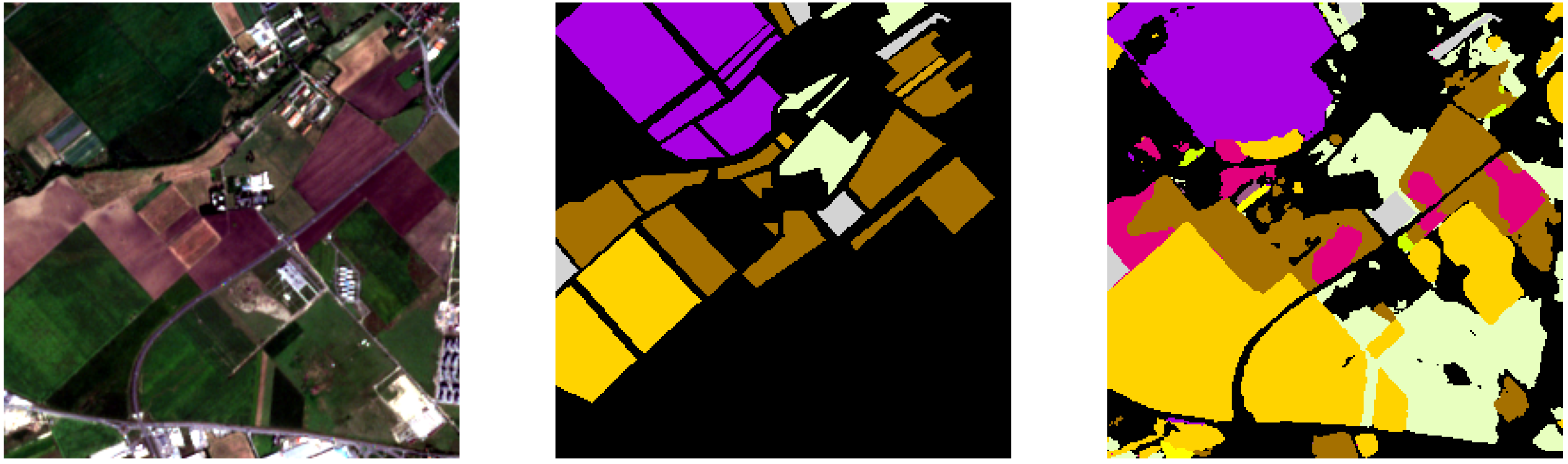}
    \caption{Qualitative prediction example on the DACIA5 April--September subset. From left to right: Sentinel-2 RGB visualization, remapped ground-truth mask, and U-Net prediction. Note that DACIA5 reference masks cover only institute-annotated parcels, so predicted agricultural regions outside the mask may correspond to unlabeled fields rather than true false positives.}
    \label{fig:dacia5_qualitative_prediction}
\end{figure}

The qualitative example in Figure~\ref{fig:dacia5_qualitative_prediction} supports the interpretation of DACIA5 as a challenging cross-dataset evaluation. The model recovers several meaningful agricultural structures, but the comparison with the reference mask must be interpreted carefully because some visible agricultural fields may be unlabeled. This supports the use of the labeled-crop-pixel protocol as the main DACIA5 evaluation setting.

The best transferred classes are maize, wheat, and pasture. These classes are both taxonomically aligned with the proposed target taxonomy and relatively well represented in the adapted PASTIS subset. In contrast, sunflower, soybean, potato, barley, and rapeseed obtain low IoU values, indicating that their discrimination is strongly affected by the change in dataset, temporal protocol, and crop appearance.

\begin{figure}[]
    \centering
    \includegraphics[width=0.95\textwidth]{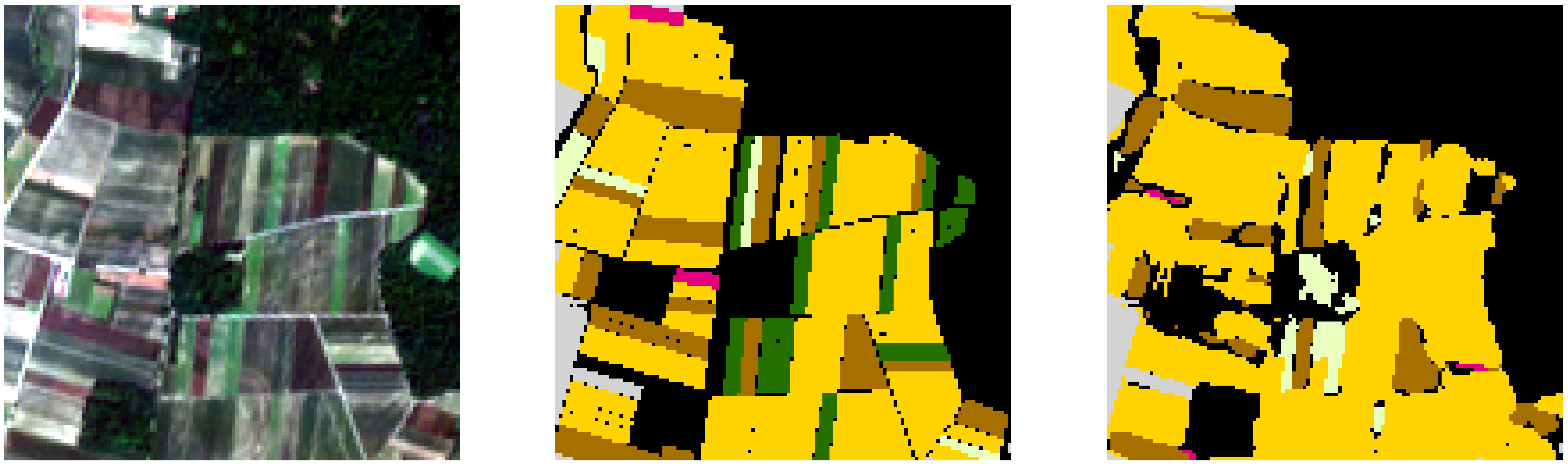}
    \caption{Qualitative prediction example on the adapted PASTIS April--September single-date subset. From left to right: Sentinel-2 RGB visualization, remapped ground-truth mask, and U-Net prediction. Class confusion remains visible due to differences in taxonomy, temporal organization, and acquisition conditions.}
    \label{fig:pastis_qualitative_prediction}
\end{figure}

The qualitative example in Figure~\ref{fig:pastis_qualitative_prediction} supports the quantitative results reported for PASTIS. Although the prediction preserves part of the spatial organization of the agricultural parcels, several crop classes are confused and some regions are over-smoothed or assigned to dominant classes. This behaviour is consistent with the lower mIoU obtained on PASTIS and highlights the difficulty of transferring a single-date model to a benchmark originally designed around Sentinel-2 time series.

The external evaluation is intended to distinguish between performance on the generated dataset and robustness under distribution shifts. A performance decrease on the external datasets is expected, especially for DACIA5 and PASTIS, because these benchmarks differ from the training data in terms of geography, label taxonomy, annotation coverage, data organization, and preprocessing assumptions. The Belgian subsets represent the least severe external shift because they still come from the EuroCrops framework, whereas DACIA5 and PASTIS represent stricter cross-dataset tests. Therefore, the results should be interpreted not only as absolute performance values, but also as indicators of how well the learned representations transfer to independent crop mapping settings.

\section{Limitations and challenges}\label{subsec:limitations_and_challenges}

Several challenges were encountered during the development of the dataset and the training process.

The internal train/validation/test split was performed at patch level. This provides a useful estimate of performance on held-out samples from the generated dataset, but it may not fully eliminate spatial correlation between nearby patches or between samples extracted from related Sentinel-2 products. To mitigate this limitation, the study includes external evaluations on unseen Belgian EuroCrops subsets, DACIA5, and PASTIS. These external tests provide a stricter assessment of model behaviour under regional and cross-dataset shifts.

First, the EuroCrops dataset exhibits inconsistencies in class definitions, naming conventions, and metadata across countries. In particular, some datasets lack complete temporal information, which complicates the alignment with Sentinel-2 acquisition periods. This required manual harmonization and filtering, introducing additional preprocessing complexity.

Second, certain crop types present high spectral similarity, especially among cereal crops such as wheat, barley, and oats. This makes their separation difficult even when using multispectral information, and may lead to confusion between classes during training.

Another important challenge is related to cloud contamination and atmospheric artefacts in Sentinel-2 imagery. Although filtering based on the Scene Classification Layer (SCL) was applied, residual cloud effects and bright artefacts may still be present in some patches, potentially affecting model performance.
 
Finally, the dataset exhibits class imbalance at both pixel and patch levels. Dominant classes such as background and major crops occupy large portions of the data, while minority classes remain sparsely represented, making the learning process more difficult for these categories.

From a modelling perspective, the limited depth of the network represents a trade-off between computational efficiency and representational capacity. While the four-level U-Net provides a good balance, more complex architectures may further improve performance, particularly for difficult class boundaries.

The Random Forest baselines also have specific limitations. The spectral Random Forest treats pixels independently and does not explicitly model parcel structure or neighbourhood relationships. As a result, its predictions may be less spatially coherent than those produced by U-Net. The spatial Random Forest partially addresses this limitation by adding local mean and standard deviation features, but these features are handcrafted and limited to fixed window sizes. Therefore, they cannot capture complex multi-scale spatial patterns in the same way as a convolutional segmentation model.

An additional limitation concerns the external evaluation on DACIA5 and PASTIS. Although these datasets are valuable for testing generalization under independent conditions, their label taxonomies and data organization do not perfectly match the reduced class system and fixed-input setting defined for the EuroCrops-based dataset. As a result, remapping and adaptation stages are required, and some categories may need to be ignored, aggregated, or temporally filtered, which can influence the interpretation of cross-dataset performance.

For DACIA5 in particular, the reference masks are based on a limited set of institute-annotated parcels rather than exhaustive annotations of all agricultural fields visible in the Sentinel-2 scenes. Therefore, background pixels in DACIA5 have a different interpretation from the background class in the EuroCrops-based dataset. This limitation motivated the use of a labeled-crop-pixel evaluation protocol as the main DACIA5 metric.

For PASTIS, an additional limitation is that the original benchmark is designed for multi-temporal analysis, whereas this work evaluates only a single-date adaptation. Furthermore, the original PASTIS arrays may contain values outside the radiometric range used by the EuroCrops-derived training patches, requiring reflectance clipping before evaluation. The spatial patch size also differs from the internally generated dataset, since PASTIS uses $128 \times 128$ patches while the proposed model was trained on $256 \times 256$ patches. Therefore, the reported PASTIS results should not be interpreted as a direct comparison with the official PASTIS benchmark protocol, but rather as a strict cross-dataset and cross-protocol transfer evaluation.

\section{Conclusions}\label{sec:conclusions}

This work presents a configurable pipeline for generating semantic-segmentation-ready crop datasets from Sentinel-2 imagery and EuroCrops parcel-level annotations, alongside a generated dataset and its evaluation using a semantic segmentation U-Net. The proposed workflow includes label harmonization, Sentinel-2 product selection, cloud and quality filtering, rasterization of parcel annotations, patch extraction, and class-aware sample selection. The resulting dataset provides aligned multispectral image--mask pairs suitable for training semantic segmentation models.

The proposed four-level U-Net model trained on the generated EuroCrops-based dataset achieved an internal test mIoU of 0.7665, a pixel accuracy of 0.8693, and a mean class accuracy of 0.9072. These results indicate that the generated data can support meaningful crop segmentation, while also showing that the selected manifest remains challenging due to class imbalance, spectral similarity between crops, and heterogeneous parcel-level annotations.

To strengthen the experimental analysis, the U-Net model was compared with Random Forest baselines. The spectral Random Forest evaluates the discriminative power of Sentinel-2 pixel-level spectral information, while the spatial Random Forest introduces handcrafted local context through mean and standard deviation features computed over fixed neighborhoods. The spatial-context Random Forest slightly improved over the spectral baseline, but both Random Forest models remained substantially below U-Net. This confirms that learned multi-scale spatial representations are important for crop segmentation from Sentinel-2 imagery.

The study also defines an external evaluation protocol based on unseen Belgian EuroCrops regions, DACIA5, and PASTIS. These datasets allow the analysis of generalization under increasingly challenging conditions, from unseen regions within the same annotation framework to cross-dataset transfer involving different taxonomies, geographic regions, and preprocessing assumptions.

External evaluation showed that the model generalizes better to unseen EuroCrops regions than to cross-dataset benchmarks with different annotation protocols. The Belgian subsets produced lower but still meaningful performance compared with the internal test split, while DACIA5 and PASTIS highlighted the difficulty of transferring a single-date EuroCrops-trained model to datasets with different label coverage, taxonomies, and temporal assumptions. The DACIA5 results showed that evaluation on labeled crop pixels is more appropriate when reference masks do not exhaustively annotate all agricultural fields visible in the Sentinel-2 scenes. PASTIS was the most difficult external setting, because it required adapting a multi-temporal benchmark to a single-date input format and also introduced differences in patch size, radiometric range, and annotation protocol.

Future work should investigate geographically disjoint training and testing protocols for a stricter assessment of spatial generalization. Additional directions include transformer-based segmentation architectures, spectral augmentation strategies, Sentinel-1 and Sentinel-2 data fusion, temporal modeling, and the integration of a two-stage land-cover and crop-specific segmentation workflow.


\section*{Declaration of Competing Interests}
The authors declare that they have no known competing financial interests or personal relationships that could have appeared to influence the work reported in this paper.

\bibliographystyle{unsrt}

\bibliography{refs}



\end{document}